%% file: main.tex
\documentclass[10pt,twocolumn,letterpaper]{article}
\PassOptionsToPackage{table}{xcolor}
\usepackage[pagenumbers]{cvpr}
\usepackage{latexml}
\input{preamble}

\iflatexml\else
\makeatletter
\patchcmd{\@maketitle}{\vskip .375in}{\vskip\dimexpr.375in-8pt\relax}
  {}{\PackageError{GAC}{Title top spacing patch failed}{}}
\patchcmd{\@maketitle}{\vspace*{24pt}}{\vspace*{20pt}}
  {}{\PackageError{GAC}{Title-author spacing patch failed}{}}
\makeatother
\fi

\title{\GACPaperTitle}
\iflatexml
\usepackage{authblk}
\author[THUHKU]{Yun~Yan\textsuperscript{*}}
\author[1]{Shizhi~Hu\textsuperscript{*}}
\author[1]{Zhichao~Li}
\author[1]{Yang~Guan\textsuperscript{\textdagger}}
\author[1]{Shiqi~Liu}
\author[1]{Jiaxin~Gao}
\author[1]{Xiangteng~Zhang}
\author[3]{Wenbo~Ding}
\author[3]{Jiuhua~Zhao}
\author[3]{Hui~Guo}
\author[2]{Xintao~Yan}
\author[1]{Keqiang~Li}
\author[1]{Shengbo~Eben~Li\textsuperscript{\textdagger}}
\affil[THUHKU]{School of Vehicle and Mobility \& College of AI, Tsinghua University\\University of Hong Kong}
\affil[1]{School of Vehicle and Mobility \& College of AI, Tsinghua University}
\affil[2]{University of Hong Kong}
\affil[3]{SAIC Intelligent Technology (Shanghai) Co., Ltd.}
\else
\author{%
{\fontsize{11}{13}\selectfont Yun Yan$^{1,2,*}$\quad Shizhi Hu$^{1,*}$\quad Zhichao Li$^{1}$\quad Yang Guan$^{1,\dagger}$\quad Shiqi Liu$^{1}$\quad Jiaxin Gao$^{1}$\quad Xiangteng Zhang$^{1}$}\\
{\fontsize{11}{13}\selectfont Wenbo Ding$^{3}$\quad Jiuhua Zhao$^{3}$\quad Hui Guo$^{3}$\quad Xintao Yan$^{2}$\quad Keqiang Li$^{1}$\quad Shengbo Eben Li$^{1,\dagger}$}\\[4pt]
\normalsize $^{1}$School of Vehicle and Mobility \& College of AI, Tsinghua University\\
\normalsize $^{2}$University of Hong Kong\\
\normalsize $^{3}$SAIC Intelligent Technology (Shanghai) Co., Ltd.\\[2pt]
{\normalsize\urlstyle{same}\url{https://yanyunn.github.io/GAC/}}
}
\fi
\date{}

\begin{document}
\iflatexml
\maketitle
\noindent\textsuperscript{*}These authors contributed equally.\quad
\textsuperscript{\textdagger}Corresponding authors.\par
\noindent\url{https://yanyunn.github.io/GAC/}\par
\input{figures/Figure1_layout}
\else
\twocolumn[{%
  \maketitle
  \vspace{-18pt}
  \input{figures/Figure1_layout}
  \par\vspace{14pt}
}]
\begingroup
\renewcommand{\thefootnote}{\fnsymbol{footnote}}
\footnotetext[1]{These authors contributed equally.}
\footnotetext[2]{Corresponding authors.}
\endgroup
\fi
\input{sections/abstract}
\input{sections/introduction}
\input{figures/Figure2_layout}
\input{sections/related_work}
\input{sections/method}
\input{sections/experiments}
\newpage % Keep surplus space below the final experimental paragraph.
\input{sections/conclusion}

\input{sections/acknowledgments}\par
{\small
\bibliographystyle{ieeenat_fullname_unsrt}
\bibliography{references}
}
\clearpage
\appendix
\makeatletter
\setlength{\@fptop}{0pt}
\setlength{\@dblfptop}{0pt}
\makeatother
\begingroup
\let\GACAddContentsLine\addcontentsline
\renewcommand{\addcontentsline}[3]{%
  \GACAddContentsLine{#1}{#2}{#3}%
  \ifstrequal{#1}{toc}{\GACAddContentsLine{atoc}{#2}{#3}}{}}
\GACAppendixContents
\input{sections/appendix}

\endgroup
\end{document}

%% file: preamble.tex
\usepackage[T1]{fontenc}
\usepackage{tabularx,array,float,listings}
\usepackage{microtype,amssymb}

\makeatletter
\renewcommand{\paragraph}{%
  \@startsection{paragraph}{4}{\z@}%
    {0.5em}{-1em}%
    {\normalfont\normalsize\bfseries}}
\makeatother

\definecolor{cvprblue}{rgb}{0.21,0.49,0.74}
\usepackage[pagebackref,breaklinks,colorlinks,allcolors=cvprblue]{hyperref}
\hypersetup{pdfauthor={Yun Yan, Shizhi Hu, Zhichao Li, Yang Guan, Shiqi Liu, Jiaxin Gao, Xiangteng Zhang, Wenbo Ding, Jiuhua Zhao, Hui Guo, Xintao Yan, Keqiang Li, Shengbo Eben Li}}
\definecolor{CaseOriginal}{HTML}{B45F5F}
\definecolor{CaseGAC}{HTML}{3F7F75}
\newcommand{\figpanelref}[3]{%
  \iflatexml Figure~\hyperref[#1]{\ref*{#1}(#3)}%
  \else Figure~\hyperref[#2]{\ref*{#1}\subref*{#2}}\fi}
\def\confName{CVPR}
\def\confYear{2027}

\newcommand{\GACPaperTitle}{Generalist Action Critic: Harnessing Frontier Models in Autonomous Driving}
\hypersetup{pdftitle={\GACPaperTitle}}
\definecolor{CallFill}{RGB}{248,248,248}
\definecolor{CallFrame}{RGB}{165,170,177}
\definecolor{GACFill}{RGB}{234,234,234}
\definecolor{HarnessPurple}{HTML}{7030A0}
\newcommand{\GACPDFRowColor}{\noalign{\begingroup\color{GACFill}\hrule height9.6pt\kern-9.6pt\endgroup}}
\newcommand{\GACHTMLRowColor}{\rowcolor{GACFill}}
\newcommand{\GACRowColor}{%
  \iflatexml\expandafter\GACHTMLRowColor
  \else\expandafter\GACPDFRowColor\fi}

\newcommand{\GACDelta}[1]{\hspace{2pt}{\color{HarnessPurple}\ensuremath{#1}}}
\newcommand{\GACDeltaBox}[2]{\makebox[#1][l]{\GACDelta{#2}}}
\newcommand{\GACScoreCell}[4][26pt]{%
  \begingroup
  \settowidth{\dimen0}{#2}%
  \ifdim\dimen0<18pt\dimen0=18pt\fi
  \settowidth{\dimen2}{#3}%
  \dimen4=\dimexpr(\dimen0-\dimen2)/2\relax
  \hspace*{\dimen4}#3%
  \GACDeltaBox{\dimexpr#1+\dimen4\relax}{#4}%
  \endgroup}
\newcommand{\GACFrequencyCell}[2]{\makebox[18pt][c]{#1}\GACDeltaBox{26pt}{#2}}
\newcommand{\GACTableStyle}{%
  \fontsize{8}{9.6}\selectfont
  \renewcommand{\arraystretch}{1.0}%
  \setlength{\extrarowheight}{0pt}}
\newcommand{\GACHTMLNumberPair}[2]{%
  \mbox{\lxAddClass{ltx_markedasmath ltx_inline-block}#1%
    \if\relax\detokenize{#2}\relax\else~\textcolor{HarnessPurple}{\ensuremath{#2}}\fi}}
\newcommand{\GACFixedCell}[4]{%
  \iflatexml\GACHTMLNumberPair{#2}{#4}%
  \else\makebox[#1][c]{#2}%
    \if\relax\detokenize{#4}\relax\hspace*{#3}\else\GACDeltaBox{#3}{#4}\fi\fi}
\newcommand{\GACFixedHeader}[2]{%
  \iflatexml\GACHTMLNumberPair{#2}{}\else\makebox[#1][c]{#2}\fi}
\newcommand{\GACMetricHeader}[3][]{%
  \iflatexml
    \if\relax\detokenize{#1}\relax\textbf{#3}%
    \else\makebox[#1][c]{\lxAddClass{ltx_inline-block}\textbf{#3}}\fi
  \else\textbf{#2}$\uparrow$\fi}
\newcommand{\GACMethodCitation}[2]{%
  \iflatexml\makebox[135pt][l]{\lxAddClass{ltx_inline-block}#1~\citep{#2}}\else#1~\citep{#2}\fi}
\newcommand{\GACAtomicHeader}[1]{%
  \iflatexml\GACHTMLNumberPair{#1}{}\else#1\fi}
\newcommand{\GACPDFTabularStar}[3]{\begin{tabular*}{#1}{#2}}
\newcommand{\GACHTMLTabularStar}[3]{\begin{tabular}{#3}}
\newcommand{\GACBeginTabularStar}[3]{%
  \iflatexml\expandafter\GACHTMLTabularStar\else\expandafter\GACPDFTabularStar\fi{#1}{#2}{#3}}
\newcommand{\GACEndPDFTabularStar}{\end{tabular*}}
\newcommand{\GACEndHTMLTabularStar}{\end{tabular}}
\newcommand{\GACEndTabularStar}{%
  \iflatexml\expandafter\GACEndHTMLTabularStar\else\expandafter\GACEndPDFTabularStar\fi}
\newcommand{\GACPDFTabular}[2]{\begin{tabular}{#1}}
\newcommand{\GACHTMLTabular}[2]{\begin{tabular}{#2}}
\newcommand{\GACBeginTabular}[2]{%
  \iflatexml\expandafter\GACHTMLTabular\else\expandafter\GACPDFTabular\fi{#1}{#2}}
\newcommand{\GACPDFRowFill}{%
  \iflatexml\else\expandafter\GACPDFRowColor\fi}
\newcommand{\GACHTMLRowFill}{%
  \iflatexml\expandafter\GACHTMLRowColor\fi}
\newcommand{\GACPDFSampleRowFill}{%
  \iflatexml\else\expandafter\GACHTMLRowColor\fi}
\iflatexml
  \renewcommand{\GACScoreCell}[4][26pt]{\GACHTMLNumberPair{#3}{#4}}
  \renewcommand{\GACFrequencyCell}[2]{\GACHTMLNumberPair{#1}{#2}}
\fi
\input{tables/TableValues}

\newcommand{\PDMSDefinition}{%
\begin{equation}
q=100\,\mathrm{NC}\cdot\mathrm{DAC}\,
\frac{5\,\mathrm{EP}+5\,\mathrm{TTC}+2\,\mathrm{C}}{12}.
\label{eq:navsim_pdms}
\end{equation}
}

\makeatletter
\newcommand{\GACAppendixContents}{%
  \begingroup
  \small
  \c@tocdepth=2\relax%
  \setlength{\parskip}{0pt}%
  \hypersetup{linkcolor=black}%
  \def\GACStripHeadingPrefix\hskip-1em.~{}%
  \renewcommand{\numberline}[1]{\hb@xt@\@tempdima{##1.\hfil}\GACStripHeadingPrefix}%
  \renewcommand{\l@section}[2]{%
    \addvspace{3pt}%
    {\bfseries\@dottedtocline{1}{0em}{1.6em}{##1}{##2}}}%
  \renewcommand{\l@subsection}[2]{%
    \@dottedtocline{2}{1.6em}{2.6em}{##1}{##2}}%
  \pdfbookmark[0]{Appendix Contents}{appendix.contents}%
  {\noindent\large\bfseries Appendix Contents\par}\smallskip
  \@starttoc{atoc}%
  \endgroup
  \par\medskip
}
\makeatother

\newcommand{\GACHTMLFullWidthGraphic}[2][]{%
  \lxAddClass{ltx_flex_figure}%
  \lxWithClass{ltx_flex_cell}{\mbox{\includegraphics[width=\textwidth,#1]{#2}}}%
  \lxWithClass{ltx_flex_break}{\mbox{}}%
}

%% file: tables/TableValues.tex
\newif\ifClosedLoopDraft
\ClosedLoopDraftfalse

\newcommand{\CLIDriveOriginal}{\GACScoreCell{\textbf{NC}$\uparrow$}{66.02}{} & \GACScoreCell{\textbf{DAC}$\uparrow$}{95.36}{} & \GACScoreCell{\textbf{TTC}$\uparrow$}{60.86}{} & \GACScoreCell{\textbf{Comfort}$\uparrow$}{98.26}{} & \GACScoreCell{\textbf{PDMS}$\uparrow$}{59.20}{} & \GACScoreCell{\textbf{RC}$\uparrow$}{48.86}{} & \GACScoreCell{\textbf{HD-Score}$\uparrow$}{39.27}{}}

\newcommand{\CLIDriveGACWithChanges}{\GACScoreCell{\textbf{NC}$\uparrow$}{71.57}{+5.55} & \GACScoreCell{\textbf{DAC}$\uparrow$}{95.47}{+0.11} & \GACScoreCell{\textbf{TTC}$\uparrow$}{67.28}{+6.42} & \GACScoreCell{\textbf{Comfort}$\uparrow$}{96.90}{-1.36} & \GACScoreCell{\textbf{PDMS}$\uparrow$}{65.20}{+6.00} & \GACScoreCell{\textbf{RC}$\uparrow$}{48.92}{+0.06} & \GACScoreCell{\textbf{HD-Score}$\uparrow$}{40.92}{+1.65}}
\newcommand{\FreqIDriveOriginal}{\GACFrequencyCell{57.58}{} & \GACFrequencyCell{51.92}{} & \GACFrequencyCell{40.78}{}}

\newcommand{\FreqIDriveOneHzWithChanges}{\GACFrequencyCell{60.68}{+3.10} & \GACFrequencyCell{51.99}{+0.07} & \GACFrequencyCell{41.59}{+0.81}}

\newcommand{\FreqIDriveTwoHzWithChanges}{\GACFrequencyCell{62.81}{+5.23} & \GACFrequencyCell{51.73}{-0.19} & \GACFrequencyCell{42.15}{+1.37}}
\newcommand{\CLDrivoROriginal}{\GACScoreCell{\textbf{NC}$\uparrow$}{54.36}{} & \GACScoreCell{\textbf{DAC}$\uparrow$}{95.96}{} & \GACScoreCell{\textbf{TTC}$\uparrow$}{48.48}{} & \GACScoreCell{\textbf{Comfort}$\uparrow$}{93.80}{} & \GACScoreCell{\textbf{PDMS}$\uparrow$}{47.56}{} & \GACScoreCell{\textbf{RC}$\uparrow$}{51.40}{} & \GACScoreCell{\textbf{HD-Score}$\uparrow$}{37.42}{}}

\newcommand{\CLDrivoRGACWithChanges}{\GACScoreCell{\textbf{NC}$\uparrow$}{64.47}{+10.11} & \GACScoreCell{\textbf{DAC}$\uparrow$}{96.20}{+0.24} & \GACScoreCell{\textbf{TTC}$\uparrow$}{58.78}{+10.30} & \GACScoreCell{\textbf{Comfort}$\uparrow$}{93.07}{-0.73} & \GACScoreCell{\textbf{PDMS}$\uparrow$}{57.61}{+10.05} & \GACScoreCell{\textbf{RC}$\uparrow$}{52.90}{+1.50} & \GACScoreCell{\textbf{HD-Score}$\uparrow$}{41.62}{+4.20}}
\newcommand{\FreqDrivoROriginal}{\GACFrequencyCell{49.36}{} & \GACFrequencyCell{52.25}{} & \GACFrequencyCell{37.99}{}}

\newcommand{\FreqDrivoROneHzWithChanges}{\GACFrequencyCell{53.67}{+4.31} & \GACFrequencyCell{54.21}{+1.96} & \GACFrequencyCell{41.04}{+3.05}}

\newcommand{\FreqDrivoRTwoHzWithChanges}{\GACFrequencyCell{59.54}{+10.18} & \GACFrequencyCell{55.17}{+2.92} & \GACFrequencyCell{43.53}{+5.54}}
\newcommand{\CLDiffusionOriginal}{\GACScoreCell{\textbf{NC}$\uparrow$}{46.26}{} & \GACScoreCell{\textbf{DAC}$\uparrow$}{93.30}{} & \GACScoreCell{\textbf{TTC}$\uparrow$}{39.74}{} & \GACScoreCell{\textbf{Comfort}$\uparrow$}{90.53}{} & \GACScoreCell{\textbf{PDMS}$\uparrow$}{37.55}{} & \GACScoreCell{\textbf{RC}$\uparrow$}{42.32}{} & \GACScoreCell{\textbf{HD-Score}$\uparrow$}{27.58}{}}

\newcommand{\CLDiffusionGACWithChanges}{\GACScoreCell{\textbf{NC}$\uparrow$}{59.43}{+13.17} & \GACScoreCell{\textbf{DAC}$\uparrow$}{94.84}{+1.54} & \GACScoreCell{\textbf{TTC}$\uparrow$}{51.99}{+12.25} & \GACScoreCell{\textbf{Comfort}$\uparrow$}{90.27}{-0.26} & \GACScoreCell{\textbf{PDMS}$\uparrow$}{50.02}{+12.47} & \GACScoreCell{\textbf{RC}$\uparrow$}{48.31}{+5.99} & \GACScoreCell{\textbf{HD-Score}$\uparrow$}{36.43}{+8.85}}
\newcommand{\FreqDiffusionOriginal}{\GACFrequencyCell{36.20}{} & \GACFrequencyCell{43.06}{} & \GACFrequencyCell{26.46}{}}

\newcommand{\FreqDiffusionOneHzWithChanges}{\GACFrequencyCell{47.26}{+11.06} & \GACFrequencyCell{46.11}{+3.05} & \GACFrequencyCell{33.00}{+6.54}}

\newcommand{\FreqDiffusionTwoHzWithChanges}{\GACFrequencyCell{50.49}{+14.29} & \GACFrequencyCell{49.26}{+6.20} & \GACFrequencyCell{36.69}{+10.23}}

%% file: figures/Figure1_layout.tex
\iflatexml
\begin{figure*}
\else
\begin{minipage}{\textwidth}
\fi
  \captionsetup{type=figure}
  \centering
  \iflatexml
  \includegraphics[width=\linewidth]{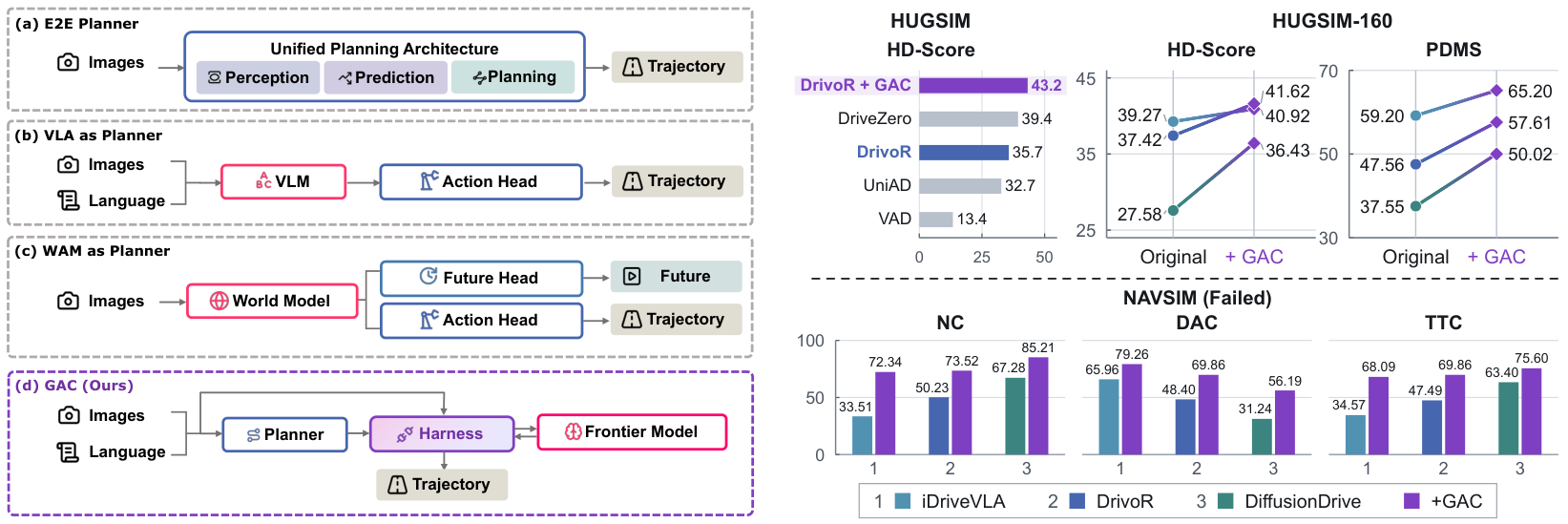}
  \else
  \includegraphics[height=58mm]{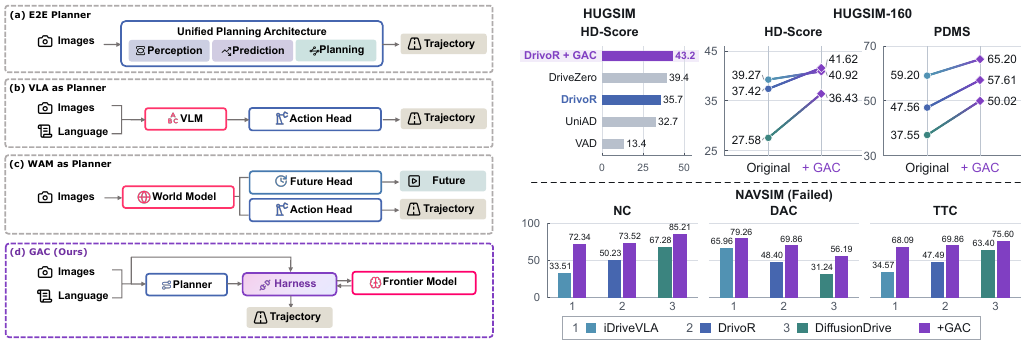}
  \fi
  \caption{\textbf{Autonomous driving paradigms and GAC results.}
  \textbf{Left: Driving paradigms.}
  (a) Planning-oriented end-to-end (E2E) driving, exemplified by UniAD~\citep{hu2023uniad}, unifies perception, prediction, and planning.
  (b) Vision-language-action (VLA) models integrate multimodal reasoning with action prediction.
  (c) World-action models (WAMs) couple predictive world representations with planning.
  (d) Generalist Action Critic (GAC) combines general-purpose frontier models with specialized driving planners through inference-time cooperation.
  \textbf{Right: Performance comparisons.}
  The first panel compares average HD-Score on the full HUGSIM benchmark, where DrivoR + GAC achieves the highest score among the compared methods.
  The remaining panels compare Original and + GAC on HUGSIM-160 (HD-Score and PDMS) and the NAVSIM v1 Failed groups (NC, DAC, and TTC).}
  \label{fig:paradigm_comparison}
\iflatexml
\end{figure*}
\else
\end{minipage}
\fi

%% file: sections/abstract.tex
\begin{abstract}
\noindent
Frontier models have demonstrated remarkable general-purpose reasoning capabilities, opening new possibilities for intelligent decision making in the physical world. In autonomous driving, many recent approaches seek to leverage these capabilities through additional adaptation for action output, particularly in scenarios where existing models underperform. We introduce \textbf{Generalist Action Critic (GAC)}, a zero-shot framework that combines frontier models' general-purpose reasoning with the planning competence of specialized driving models through an inference-time harness. In GAC, specialized planners generate multiple candidate trajectories and make an initial selection, while a frozen frontier model serves as an action critic that assesses the selected trajectory against scene evidence and either retains it or reselects from the available candidates. The harness combines camera images with spatial overlays, provides driving context and reasoning principles through textual instructions, and regulates the information available at each stage to support trajectory assessment and reselection. On HUGSIM, GAC raises DrivoR's HD-Score to 43.2, achieving state-of-the-art performance among methods trained without additional simulation data. GAC also consistently improves closed-loop driving safety across three planners, with relative HD-Score gains of up to 32.1\%. Experiments on NAVSIM v1 further reveal how GAC improves planning performance by correcting up to 52.13\% of eligible unsafe trajectory selections while introducing new failures in only 0.67\% of originally high-quality selections. These results demonstrate the potential of harnessing frontier models to improve driving safety through inference-time cooperation with specialized planners.
\end{abstract}

%% file: sections/introduction.tex
\section{Introduction}
\label{sec:introduction}

Recent advances in artificial intelligence have significantly expanded the capabilities of general-purpose models. From language understanding and generation to multimodal reasoning, complex problem solving, and tool use, frontier models are breaking traditional boundaries in the range of tasks that general-purpose AI can perform~\citep{openai2026astra,anthropic2026fable}. This progress has now reached the physical world, with recent studies demonstrating GPT-6 Astra's strong spatial reasoning and decision-making capabilities in robotic manipulation, navigation, and other embodied AI tasks~\citep{unexpected_robot_policy_2026,astra_vln_2026}.

Autonomous driving is a representative domain of embodied AI, requiring systems to connect scene understanding with driving decisions and vehicle control~\citep{wu2022tcp,guan2022integrated,hu2023uniad,guan2026enhanced}. Existing approaches organize these capabilities through different architectures, including planning-oriented end-to-end (E2E)~\citep{hu2022stp3,hu2023uniad,jiang2023vad,liao2025diffusiondrive}, vision-language-action (VLA)~\citep{hwang2025emma,renz2025simlingo,li2026recogdrive,nvidia2025alpamayo}, and world--action models (WAM)~\citep{zheng2025world4drive,zhou2026drivedreamerpolicy}, as illustrated in Figure~\ref{fig:paradigm_comparison}. Despite their different architectures, these paradigms have largely advanced through task-specific training and adaptation. As frontier models continue to advance, a key question is how to directly leverage their general-purpose reasoning capabilities to improve driving decisions.

Multi-candidate planning provides a concrete setting for exploring this question. The inherently multimodal nature of driving behavior has motivated a growing shift from predicting a single trajectory toward generating multiple candidate trajectories~\citep{li2024hydra,li2025gtrs}. Effective candidate selection is therefore essential to realizing the full potential of multimodal planning. Yet existing planners do not always select the best available trajectories, leaving a gap between candidate generation and selection~\citep{he2026idrivevla}. Frontier models' strong spatial reasoning offers a promising way to address this gap.

To this end, we introduce \textbf{Generalist Action Critic (GAC)}, a new paradigm for autonomous driving that combines general-purpose frontier models with specialized driving planners through inference-time cooperation. Rather than adapting frontier models for driving action output, GAC directly employs them as action critics to assess and improve the decisions of specialized driving systems. In this work, we instantiate GAC for multi-candidate trajectory planning, where the critic evaluates the planner's initial selection and reselects from the available candidates when needed. Effective trajectory assessment requires understanding how candidate trajectories interact with the surrounding environment and determining whether they are safe and appropriate for the driving situation. We therefore develop an evidence-grounded inference harness that integrates layered visual representations with a structured reasoning context to support selective decision revision. We envision GAC as a step toward a new generation of autonomous driving paradigms, where advances in general-purpose intelligence and specialized planning can jointly shape future driving systems.

Our main contributions are summarized as follows:

\begin{itemize}
    \item We introduce Generalist Action Critic (GAC), a new paradigm for autonomous driving that enables zero-shot, inference-time cooperation between general-purpose frontier models and specialized driving planners.

    \item We develop an evidence-grounded inference harness that integrates layered visual representations, a structured reasoning context, and selective decision revision to support trajectory evaluation based on scene evidence.

    \item We demonstrate GAC's effectiveness across three driving planners on HUGSIM and NAVSIM v1. GAC improves HD-Score by up to 32.1\%, increases DrivoR's HD-Score to 43.2 on the full HUGSIM benchmark, and recovers up to 52.13\% of eligible zero-score failures.
\end{itemize}

%% file: figures/Figure2_layout.tex
\begin{figure*}[t]
    \centering
\iflatexml
    \includegraphics[width=\textwidth]{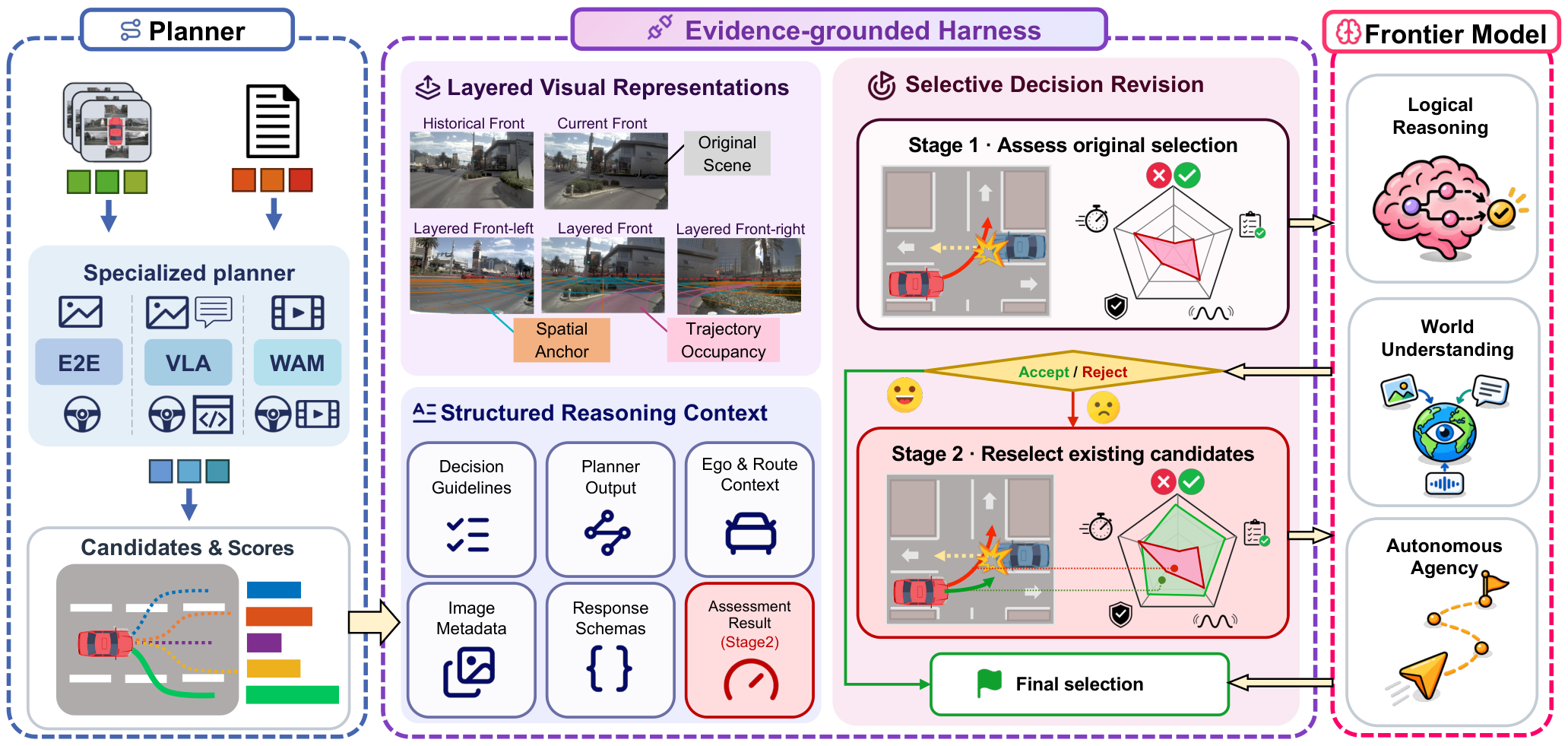}
\else
    \includegraphics[width=\textwidth]{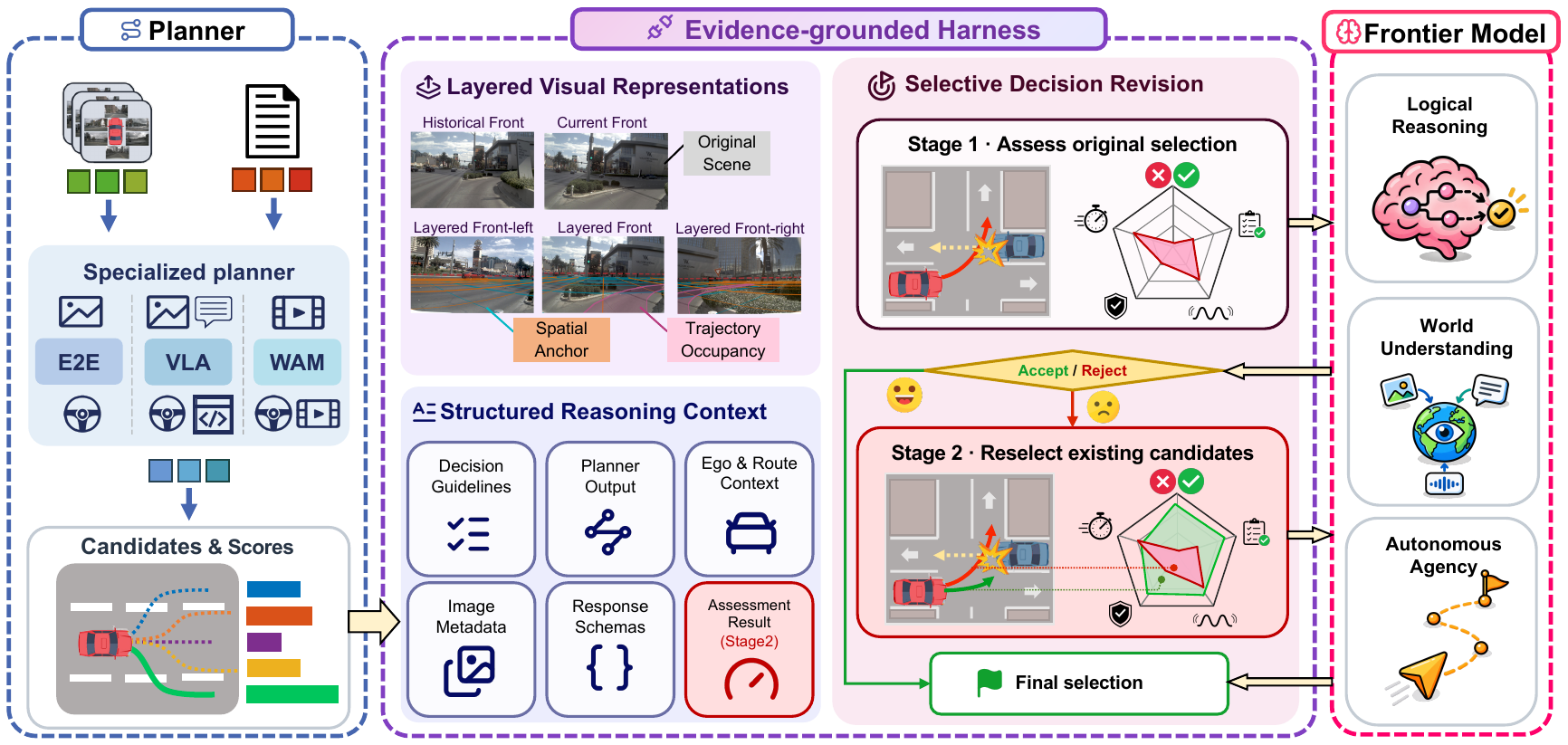}
\fi%
    \caption{Overview of Generalist Action Critic (GAC). A specialized planner generates candidate trajectories and provides its original selection. An evidence-grounded inference harness integrates layered visual representations and a structured reasoning context to support a frontier model serving as the action critic. Through selective decision revision, the critic assesses the original selection, retaining it upon acceptance and reselecting from the same candidate set upon rejection.}
    \label{fig:framework}
\end{figure*}

%% file: sections/related_work.tex
\section{Related Work}\label{sec:related_work}
\paragraph{General-purpose models for embodied reasoning.}
Research on embodied intelligence has explored how general-purpose models can support reasoning and decision making in the physical world. Early work connected language models with executable robot skills, environmental feedback, spatial representations, and visual annotations for action selection~\citep{ichter2023saycan,huang2023innermonologue,huang2023voxposer,nasiriany2024pivot}. In autonomous driving, language and vision-language models have been used for driving decisions, reasoning-assisted planning, and cooperation with conventional planning systems~\citep{dilu,llm_assist,tian2024drivevlm}. However, earlier models faced limitations in spatial reasoning and reliable physical execution, constraining their roles in complex embodied tasks~\citep{Chen_2024_CVPR,pmlr-v267-yang25f}. Recent advances in frontier models, particularly GPT-6 Astra, have demonstrated stronger capabilities in robotic manipulation and expanded their roles in direct action generation and cooperation with learned policies~\citep{unexpected_robot_policy_2026,astra_hybrid_policy_report}. Frontier models have also been evaluated in closed-loop driving~\citep{drivingbench_2026}. These advances motivate our exploration of frontier models as action critics for evaluating and improving driving decisions.

\paragraph{Data-driven autonomous driving.}\label{sec:related_planning}
Learning-based approaches have explored diverse architectures for driving perception, planning, and control~\citep{hu2022stp3,guan2022integrated,wu2022tcp,shao2023interfuser,guan2026enhanced}. UniAD~\citep{hu2023uniad} exemplifies a unified, planning-oriented end-to-end architecture that integrates perception, prediction, and planning. Subsequent studies have advanced scene representations, planning efficiency, and the interaction between prediction and planning~\citep{jiang2023vad,weng2024paradrive,chen2024ppad}. Vision-language approaches incorporate pretrained multimodal knowledge and reasoning into driving, supporting scene understanding, language-guided control, and trajectory planning~\citep{shao2024lmdrive,sima2024drivelm,hwang2025emma,renz2025simlingo,nvidia2025alpamayo}. World-model-based approaches connect predictive world representations with action planning, allowing predicted scene dynamics to inform driving decisions~\citep{zheng2025world4drive,zhang2026identifiabilitycontrolledworldmodels,wang2026latentwam,zhang2026worldactioncriticwac,zhou2026drivedreamerpolicy}. GAC builds on these architectures by introducing a new paradigm for inference-time cooperation between frontier models and specialized driving planners.

\paragraph{Multimodal planning and trajectory selection.}
Earlier planning approaches primarily predicted a single ego trajectory, while recent methods increasingly model multiple possible driving behaviors to capture the multimodal nature of driving~\citep{li2024hydra,jiang2026vadv2,guan2027beyond,kirby2026drivor}. Candidate sets are constructed using predefined trajectory vocabularies~\citep{jiang2026vadv2}, scene-conditioned proposals~\citep{guo2025ipad,kirby2026drivor}, and generative trajectory models~\citep{zheng2024genad,liao2025diffusiondrive}.

Selecting the trajectory to execute is therefore a critical component of multimodal planning. Hydra-MDP and DrivoR learn to predict simulation-derived planning scores for candidate evaluation~\citep{li2024hydra,kirby2026drivor}. GTRS improves the generalization of learned trajectory scoring by training on a dense trajectory vocabulary with trajectory dropout~\citep{li2025gtrs}. More recently, iDriveVLA identifies the gap between candidate quality and selection performance, introducing a safety-aware scorer and a trained modulator that adjusts criterion weights using frozen VLM features~\citep{he2026idrivevla}. GAC instead leverages frontier models' general-purpose reasoning for trajectory assessment and reselection without additional training.

%% file: sections/method.tex
\section{Generalist Action Critic}\label{sec:method}
\subsection{Framework}\label{sec:framework}
\input{sections/framework}

\subsection{Layered Visual Representations}\label{sec:images}

\begin{figure*}[t]
\input{figures/Figure3_layout}
\end{figure*}
Trajectories are expressed in ego coordinates, whereas actors and road boundaries are observed through camera images. To bridge this gap, we develop layered visual representations that combine spatiotemporal scene context $V$, spatial anchor projection $A$, and future occupancy projection $O$, as illustrated in Figure~\ref{fig:geometry_layers}. Together, these layers provide visual evidence of the driving scene, metric spatial references, and the projected vehicle occupancy along candidate motion.

\paragraph{Spatiotemporal scene context ($V$).}
Temporal context helps the critic infer nearby actor motion, while multiple viewpoints reveal surrounding road geometry and actors during turns. We use current front, front-left, and front-right camera images, together with a historical front image from $1.5$ seconds earlier. The critic interprets changes between the historical and current front images alongside ego motion to infer qualitative motion trends. The five-image input is
\begin{equation}
I=\left[I_{\mathrm F}^{\mathrm h},I_{\mathrm F},
\widetilde I_{\mathrm F},\widetilde I_{\mathrm L},\widetilde I_{\mathrm R}\right],
\label{eq:visual_attachments}
\end{equation}
where F, L, and R denote the front, front-left, and front-right cameras, respectively. The superscript h denotes the historical front image, and the tilde indicates images augmented with spatial anchors $A$ and future occupancy $O$. The historical and current front images preserve scene details without annotation overlays, while the three annotated views relate the candidate's projected occupancy to surrounding actors and road boundaries.

\paragraph{Spatial anchor projection ($A$).}
Perspective makes image distance a poor proxy for metric clearance. We therefore construct spatial anchors as a metric grid on an assumed flat ground plane in the same fixed ego coordinate frame as the candidate trajectories. The anchors cover an $80\,\mathrm m\times80\,\mathrm m$ region at $5$-metre intervals, providing metric references for estimating distances and lateral clearance while limiting dense markings near the horizon.

We project these anchors into the front, front-left, and front-right camera views using calibrated camera intrinsics and extrinsics. The projection also includes the theoretical horizon, corresponding to viewing rays parallel to the assumed ground plane. Calibrated lens-distortion matching aligns the projected annotations with the raw camera images when necessary. 

\paragraph{Future occupancy projection ($O$).}
Candidate trajectories describe the motion of a vehicle reference point, while assessing potential collisions and road-boundary violations requires considering the full vehicle footprint. We therefore construct the union of vehicle footprints along the planner's originally selected trajectory $\tau_b$. The vehicle footprint is represented by $B=[-\ell_r,\ell_f]\times[-w/2,w/2]$, where $\ell_r$ and $\ell_f$ are the distances from the trajectory reference point to the rear and front edges, and $w$ is the vehicle width. The resulting swept occupancy is defined as
\begin{equation}
\mathcal U=
\bigcup_{t\in\{0,0.02,\ldots,4.0\}\,\mathrm{s}}
\left\{\mathbf R(\psi_b^t)\mathbf p+\mathbf q_b^t:\mathbf p\in B\right\}.
\label{eq:critic_swept_footprint}
\end{equation}
Here, $\mathbf R(\psi)$ is a two-dimensional rotation matrix, and $\mathbf q_b^t=(x_b^t,y_b^t)^\top$ denotes the reference-point position in the fixed current ego frame. Positions and unwrapped headings are linearly interpolated from the initial pose and eight future samples to obtain poses at $0.02$-second intervals.

We project the resulting footprint union $\mathcal U$ into the camera images using the same calibrated mapping as the spatial anchors, as illustrated in \figpanelref{fig:geometry_layers}{fig:geometry_input}{b}. Lens-distortion matching resamples only the annotation layers, preserving the original camera images. Further implementation details are provided in Appendix~\ref{app:implementation}.

\input{tables/Table1}
\subsection{Structured Reasoning Context}\label{sec:context}
The harness constructs a structured reasoning context for each stage by organizing candidate trajectories, ego state, navigation information, image metadata, decision guidelines, and response requirements as textual inputs. The reselection stage additionally receives the assessment result from the first stage. These stage-specific contexts are defined as follows:
\begin{equation}
\begin{aligned}
T_1&=\bigl(G_1,P_1,E,M,S_1\bigr),\\
T_2&=\bigl(G_2,P_2,E,M,S_2,R\bigr).
\end{aligned}
\label{eq:text_layer}
\end{equation}
Here, subscripts $1$ and $2$ denote the assessment and reselection stages, respectively. $G_1$ and $G_2$ denote the corresponding decision guidelines. The candidate inputs are $P_1=(\{\tau_b\},b)$ and $P_2=(\mathcal C,b)$, containing the original selection and the full candidate set, respectively, together with the original selection index $b$. The shared context $E$ supplies ego pose, velocity, acceleration, and route intent, while image metadata $M$ identify camera views, observation times, and annotations. The response schema $S_1$ specifies an acceptance decision and its supporting reason, while $S_2$ specifies a selected candidate index and a comparative reason. The assessment result $R=(a,r_1)$ contains the acceptance decision and its reason.

\paragraph{Decision guidelines ($G_1,G_2$).}
The guidelines establish shared conventions for coordinate frames, timestamps, ego state, route information, and image annotations. They also specify the evaluation metrics and their aggregation into an overall driving score. Building on these definitions and conventions, we specify three shared reasoning principles:

\begin{enumerate}[label=\textit{(\arabic*)},leftmargin=*,labelsep=0.5em,topsep=0pt,partopsep=0pt,itemsep=2pt,parsep=0pt]
\item \textit{Spatiotemporal reasoning.}
The guidelines require reasoning about the ego vehicle's position relative to actors and road boundaries over time, considering its full footprint. For potential conflicts with other actors, the critic combines time-indexed candidate poses with historical and current observations to assess whether the ego vehicle and other actors may occupy overlapping regions at the same time.

\item \textit{Evidence and uncertainty.}
The guidelines require explicit evidence for every decision, including acceptance. Justifications must explain the expected consequences of candidate motion based on scene observations and specify their location and approximate timing where relevant. These explanations must distinguish observed evidence from assumptions about future behavior and explicitly acknowledge uncertainty when it affects the judgment.

\item \textit{Performance tradeoffs.}
The guidelines require the critic to weigh safety, progress, and comfort when evaluating candidate motion, while accounting for traffic rules. Improvements in one aspect must be considered alongside potential costs in others. Lower speed or shorter travel distance alone does not establish better performance.
\end{enumerate}

\noindent Both stages share these reasoning principles but require different justifications. Assessment ($G_1$) requires evidence that supports acceptance or substantiates a specific concern warranting rejection. Reselection ($G_2$) requires comparative evidence explaining why the final candidate is preferred.

\subsection{Selective Decision Revision}\label{sec:revision}
GAC performs selective decision revision using layered visual representations and a structured reasoning context. The critic first assesses the planner's original selection and compares alternative candidates only upon rejection. Each stage uses its own model context with complete instructions.

In the first stage, the critic evaluates the planner's selected trajectory $\tau_b$ based on the driving context and decision guidelines. Given the shared visual input $I$ and assessment input $T_1$, the critic returns an acceptance decision $a\in\{0,1\}$ and a supporting reason $r_1$:
\begin{equation}
(a,r_1)=F(I,T_1).
\label{eq:critic_assessment}
\end{equation}
Here, $a=1$ indicates acceptance, while $a=0$ triggers reselection. If accepted, the planner's original trajectory is retained without further candidate comparison.

Upon rejection, the critic uses $T_2$ and the shared visual input $I$ to reassess the concern identified by $r_1$ and compare candidates in the full set $\mathcal C$ based on expected driving performance. The visual input remains unchanged, with its overlays depicting the planner's original selection $\tau_b$. The critic returns the preferred candidate index $j$ and a comparative reason $r_2$:
\begin{equation}
(j,r_2)=F(I,T_2),\qquad j\in\{0,\ldots,K-1\}.
\label{eq:critic_reselection}
\end{equation}
The original trajectory remains eligible for reselection and may be selected again if the initial concern is unsupported or no alternative offers better expected performance. The final trajectory is determined by
\begin{equation}
\tau_c =
\begin{cases}
\tau_b, & a=1,\\
\tau_j, & a=0.
\end{cases}
\label{eq:final_trajectory}
\end{equation}

%% file: sections/framework.tex
The Generalist Action Critic (GAC) framework consists of three components: a specialized driving planner, an evidence-grounded inference harness, and a general-purpose frontier model serving as the action critic, as illustrated in Figure~\ref{fig:framework}. Given a driving scene, the planner generates and scores $K$ candidate trajectories $\mathcal C=\{\tau_0,\ldots,\tau_{K-1}\}$ and selects $\tau_b$, where $b$ denotes the original selection index. These outputs are passed to the harness through a common trajectory interface that supports different planning architectures. The harness integrates layered visual representations with a structured reasoning context, providing shared visual input $I$ and stage-specific textual inputs $T_1$ and $T_2$ to the action critic $F$. The critic performs selective decision revision by assessing the original selection and, upon rejection, reselecting from the existing candidates. Both the planner and the frontier model remain frozen.

Each candidate trajectory is represented by eight time-indexed poses over a 4-second horizon:
\begin{equation}
\tau_i=\left\{(x_i^t,y_i^t,\psi_i^t)\right\}_{t\in\{0.5,1.0,\ldots,4.0\}\,\mathrm{s}}.
\label{eq:timed_candidate}
\end{equation}
Here, $x_i^t,y_i^t$ denote forward and leftward positions in metres in a fixed ego coordinate frame at the current time, and $\psi_i^t$ denotes heading relative to the current ego orientation in radians. The initial pose $(0,0,0)$ at $t=0$ is separate from these future samples.

%% file: figures/Figure3_layout.tex
\centering
\iflatexml\else
\begingroup%
\phantomsubcaption\label{fig:geometry_construction}%
\phantomsubcaption\label{fig:geometry_input}%
\endgroup%
\fi%
\iflatexml
\includegraphics[width=\linewidth,alt={(a) Geometric construction and (b) Five-image input. Spatial anchors and future vehicle occupancy are projected into camera views.}]{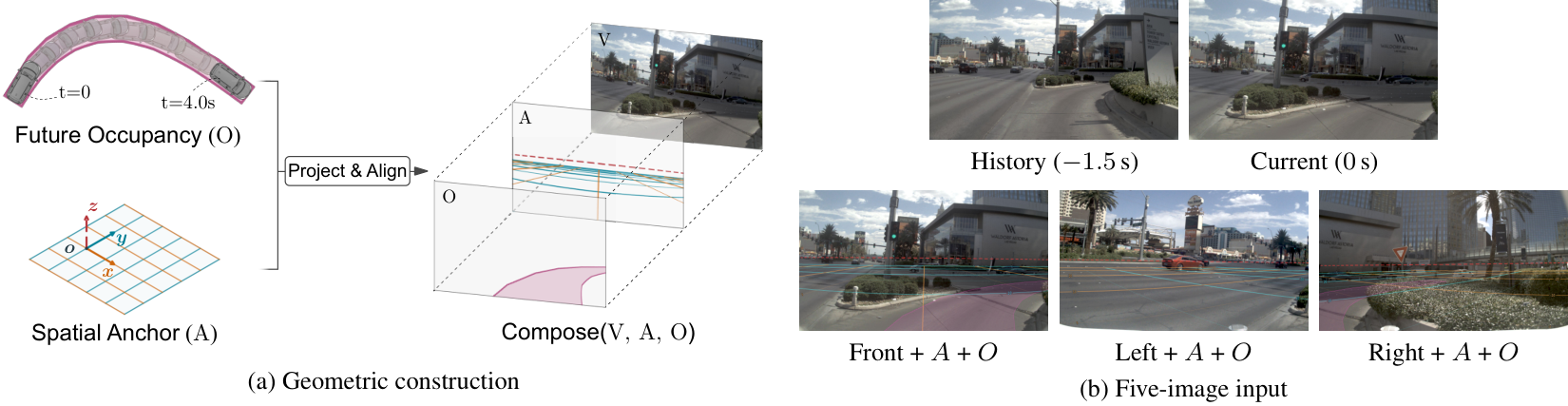}%
\else
\includegraphics[width=\linewidth,alt={(a) Geometric construction and (b) Five-image input. Spatial anchors and future vehicle occupancy are projected into camera views.}]{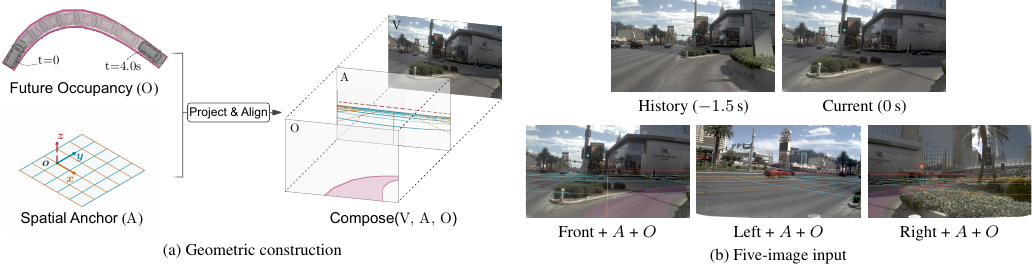}%
\fi%
\caption{Layered visual representations for the action critic. (a) Spatial anchors and vehicle footprints are projected into camera views with lens distortion accounted for. (b) The five-image input consists of historical and current front views and three annotated current views. The overlays combine spatial anchors ($A$) with the original candidate's future vehicle occupancy ($O$).}
\label{fig:geometry_layers}

%% file: tables/Table1.tex
\begin{table*}[t]
\centering
\caption{Closed-loop performance on the full HUGSIM benchmark (345 scenarios) without additional simulation training data. E/M/H/X denote Easy/Medium/Hard/Extreme. Bold and underlined values indicate the best and second-best results per column, respectively.}
\label{tab:closed_loop_full}
\GACTableStyle
\setlength{\tabcolsep}{10.5pt}
\begin{tabular}{@{}l@{\hspace{14pt}}*{4}{c}|c@{\hspace{26pt}}*{4}{c}|c@{}}
\toprule
\raisebox{-6pt}[0pt][0pt]{\textbf{Method}} & \multicolumn{5}{@{}c@{\hspace{26pt}}}{\textbf{RC}$\uparrow$} & \multicolumn{5}{@{}c@{}}{\textbf{HD-Score}$\uparrow$}\\
\cmidrule(r{26pt}){2-6}\cmidrule{7-11}
 & \textbf{E} & \textbf{M} & \textbf{H} & \textbf{X} & \textbf{Avg.} & \textbf{E} & \textbf{M} & \textbf{H} & \textbf{X} & \textbf{Avg.}\\
\midrule
LTF~\citep{chitta2022transfuser} & 67.8 & 35.1 & 26.2 & 40.5 & 38.9 & 58.9 & 18.0 & 9.8 & 25.9 & 23.7\\
VAD~\citep{jiang2023vad} & 51.3 & 31.1 & 25.3 & 26.5 & 31.4 & 36.3 & 9.5 & 8.0 & 11.5 & 13.4\\
UniAD~\citep{hu2023uniad} & 78.4 & 60.5 & 33.6 & 17.8 & 45.9 & 64.9 & 45.8 & \underline{20.6} & 6.6 & 32.7\\
GigaPixel~\citep{rowe2026gigapixel} & 78.6 & 60.7 & 32.7 & 35.4 & 50.1 & 67.4 & \textbf{51.9} & 19.1 & 21.6 & 38.5\\
DriveZero~\citep{xiaomi2026drivezero} & \textbf{85.5} & \textbf{63.7} & 31.7 & 43.3 & \underline{53.9} & \textbf{78.5} & 47.2 & 16.0 & 28.7 & \underline{39.4}\\
DrivoR~\citep{kirby2026drivor} & 80.9 & 50.5 & \underline{33.8} & \textbf{47.1} & 49.8 & 73.3 & 34.6 & 18.8 & \underline{32.5} & 35.7\\
\GACRowColor
DrivoR + GAC & \underline{82.8} & \underline{63.4} & \textbf{35.8} & \underline{46.1} & \textbf{55.0} & \underline{75.8} & \underline{50.3} & \textbf{24.1} & \textbf{33.5} & \textbf{43.2}\\
\bottomrule
\end{tabular}
\par\vspace{-3pt}% Local adjustment of the table-to-text gap.
\end{table*}

%% file: sections/experiments.tex
\section{Experiments}\label{sec:evaluation}
\setcounter{dbltopnumber}{2}
\input{tables/Table2}
\input{figures/Figure4_layout}
\input{tables/Table3}
\input{tables/Table4}

\subsection{Experimental Setup}\label{sec:setup}\label{sec:protocols}
\paragraph{Planners and frontier model.}
We evaluate GAC with iDriveVLA, DrivoR, and DiffusionDrive~\citep{he2026idrivevla,kirby2026drivor,liao2025diffusiondrive}, which generate 64, 64, and 20 candidate trajectories, respectively. Each planner is paired with GPT-6 Astra~\citep{openai2026astra} at maximum reasoning effort. Original denotes the planner's native selection, while GAC assesses and reselects from the same candidate set. Candidate evaluation scores and privileged evaluator information are withheld from the critic.

\paragraph{Closed-loop evaluation.}
We evaluate GAC on HUGSIM~\citep{zhou2024hugsim}, using its 345-scenario evaluation set for comparison with existing methods. To enable cross-planner evaluation while keeping frontier-model usage costs manageable, we use HUGSIM-160, a balanced set of 160 HUGSIM scenarios with ten scenarios from each combination of four data sources (KITTI-360, nuScenes, PandaSet, and Waymo) and four difficulty levels (Easy, Medium, Hard, and Extreme).

For each planner, Original and GAC share the same checkpoints, sensor adaptations, controllers, and simulation settings. Planning runs at 4\,Hz, with GAC assessment at 2\,Hz unless otherwise specified. The simulator waits for critic decisions before advancing.

\paragraph{Open-loop evaluation.}
We export candidate trajectories for all 12,146 scenes in the NAVSIM v1 navtest split~\citep{dauner2024navsim}. Based on the original selection's PDMS, we divide scenes into Failed ($\mathrm{PDMS}=0$), Intermediate ($0<\mathrm{PDMS}<90$), and High-quality ($\mathrm{PDMS}\geq90$). The Failed group includes zero-score selections with at least one positive-score alternative, comprising 188, 219, and 541 eligible scenes for iDriveVLA, DrivoR, and DiffusionDrive, respectively. For the Intermediate and High-quality groups, we sample 100 distinct scenes three times per planner and average the results across the three samples.

\paragraph{Evaluation metrics.}\label{sec:pdms}\label{sec:metrics}
We follow the official HUGSIM and NAVSIM v1 evaluation protocols. Both benchmarks use the PDM Score (PDMS), which combines a collision-related metric (NC), drivable area compliance (DAC), time-to-collision within bound (TTC), and comfort (C), with NC and DAC serving as multiplicative factors. NAVSIM additionally includes ego progress (EP). On HUGSIM, route completion (RC) measures progress along the prescribed route, and HD-Score is the product of RC and the mean PDMS over an episode. For the NAVSIM ablations, we additionally report the per-instance $\mathrm{NC}\times\mathrm{DAC}$ score to quantify collision avoidance and drivable-area compliance jointly. All component scores are normalized to $[0,1]$, and reported results are scaled to 0--100, with higher values indicating better performance. Additional evaluation details are provided in Appendix~\ref{app:experimental_details}.

\subsection{Closed-Loop Evaluation}\label{sec:closed_loop}
\paragraph{Overall performance.}
We evaluate DrivoR with GAC on the full HUGSIM benchmark. As shown in Table~\ref{tab:closed_loop_full}, GAC improves DrivoR's HD-Score from 35.7 to 43.2 (+21.0\%) and RC from 49.8 to 55.0, achieving state-of-the-art performance among the compared methods trained without additional simulation data.

\paragraph{Cross-planner performance.}
We further evaluate GAC across three planners on HUGSIM-160, as shown in Table~\ref{tab:closed_loop_results}. GAC consistently improves NC and TTC across all three planners, demonstrating improved driving safety alongside gains in PDMS and HD-Score. DiffusionDrive achieves the largest relative HD-Score gain of 32.1\%. These improvements are accompanied by stable or higher RC. Notably, GAC further improves iDriveVLA despite its existing safety-aware scoring mechanism.

\paragraph{Performance across difficulty levels.}
Figure~\ref{fig:closed_loop_difficulty} shows that GAC improves HD-Score and NC across all four difficulty levels for all three planners. The largest HD-Score gains occur in Medium scenarios for each planner. The smaller gains in Easy and Extreme scenarios may reflect the strong baseline performance in Easy scenarios and the persistent difficulty of completing routes in Extreme scenarios. We provide detailed breakdowns in Appendix~\ref{app:closed_loop}.

\paragraph{Assessment frequency.}
We evaluate GAC at 1\,Hz and 2\,Hz on HUGSIM-64, a balanced subset of HUGSIM-160 containing four scenarios per source--difficulty combination. Planning remains fixed at 4\,Hz. As shown in Table~\ref{tab:frequency}, both assessment frequencies improve PDMS and HD-Score over native selection across all three planners, with 2\,Hz yielding larger gains in both metrics.

\subsection{Open-Loop Selection Analysis}\label{sec:sampled}\label{sec:intervention}
Closed-loop evaluation measures driving performance over complete episodes, while open-loop evaluation examines individual trajectory selections. We analyze failure recovery and decision preservation on NAVSIM v1.

\paragraph{Failure recovery.}
Table~\ref{tab:combined_components} presents the results across the three groups. In the Failed groups, GAC improves NC, DAC, and TTC across all three planners, recovering 42.51--52.13\% of eligible zero-score selections. Mean PDMS increases from zero to 36.26--47.41, demonstrating that improved trajectory selection can recover decisions that originally failed safety-related evaluation.

\paragraph{Improvement and preservation.}
Beyond recovering failed selections, GAC also improves mean PDMS across all three planners in the Intermediate groups, with consistent gains in TTC. For High-quality selections, NC, DAC, and TTC remain at or above 99.00, while only six new zero-score failures occur among 900 evaluated instances, yielding an overall new failure rate of 0.67\%. Mean PDMS decreases by 1.46--2.84 points, largely due to reduced ego progress, reflecting more conservative trajectory selection. Further analysis of the safety-degradation cases shows that violations first occur after 1\,s of the four-second evaluation horizon. In closed-loop evaluation, GAC reassesses selections every 0.5\,s using updated observations, allowing decisions to be revised before such risks materialize. Further analyses are provided in Appendix~\ref{app:additional_results}.

\input{figures/Figure5_layout}
\subsection{Ablation Studies}\label{sec:ablations}

We examine the contributions of selective decision revision and its visual and contextual inputs using matched iDriveVLA candidate sets on NAVSIM.

\paragraph{Workflow ablations.}
Table~\hyperref[tab:harness_ablations]{\ref*{tab:harness_ablations}(a)} compares GAC with three alternative workflows. Uniform retains GAC's assessment decisions and samples uniformly from the candidate set upon rejection. Direct selects from all candidates without prior assessment, while Select then review evaluates a newly selected candidate against the original choice. GAC substantially outperforms Uniform on Failed-group PDMS, demonstrating the importance of evidence-based candidate comparison. Although Direct achieves higher Failed-group PDMS, GAC better preserves the safety of High-quality selections, achieving an NC$\times$DAC score of 99.00 versus 91.00. When both methods are evaluated on the same 54 failed cases identified by GAC's assessment stage, GAC recovers 50 compared with 44 under Direct, demonstrating the benefit of assessment-guided reselection. Select then review also yields lower PDMS than GAC in both groups.

\input{tables/Table5}
\paragraph{Input ablations.}
Table~\hyperref[tab:harness_ablations]{\ref*{tab:harness_ablations}(b)} examines the contributions of reasoning principles and geometric overlays. Failed-group PDMS drops from 49.26 to 29.22 without reasoning principles and to 27.42 without geometric overlays. Geometric overlays also help preserve the safety of High-quality selections, improving NC$\times$DAC from 97.00 to 99.00. Ablation details are provided in Appendix~\ref{app:ablation_configurations}.

\subsection{Qualitative Analysis}\label{sec:cases}
\figpanelref{fig:qualitative}{fig:qualitative_open}{a} illustrates open-loop recovery. GAC selects a trajectory that delays the right turn to reduce a potential pedestrian conflict at 2.5--3.5\,s. On the narrow bend, the original trajectory's projected footprint extends beyond the curb, so GAC selects a candidate with greater clearance. PDMS increases from zero to 95.10 and 100, respectively. \figpanelref{fig:qualitative}{fig:qualitative_closed}{b} shows closed-loop driving behavior. With DrivoR (top), GAC reduces the closing speed before passing the vehicle ahead. With DiffusionDrive (bottom), it selects an alternative that stays farther right to maintain clearance from a front-facing vehicle. In both closed-loop cases, Original collides, whereas GAC completes the route. See Appendix~\ref{app:qualitative} for additional examples.

%% file: tables/Table2.tex
\begin{table*}[t]
\centering
\caption{Closed-loop results for three planners on HUGSIM-160. Shading marks GAC; purple values show score changes from Original.}
\label{tab:closed_loop_results}
\GACTableStyle
\setlength{\tabcolsep}{4pt}
\GACBeginTabularStar{\textwidth}{@{}>{\raggedright\arraybackslash}m{91pt}@{\extracolsep{\fill}}*{7}{l}@{}}{@{}l*{7}{c}@{}}
\toprule
\textbf{Method} & \GACScoreCell{\GACMetricHeader{NC}{NC~↑}}{\GACMetricHeader{NC}{NC~↑}}{} & \GACScoreCell{\GACMetricHeader{DAC}{DAC~↑}}{\GACMetricHeader{DAC}{DAC~↑}}{} & \GACScoreCell{\GACMetricHeader{TTC}{TTC~↑}}{\GACMetricHeader{TTC}{TTC~↑}}{} & \GACScoreCell{\GACMetricHeader{Comfort}{Comfort~↑}}{\GACMetricHeader{Comfort}{Comfort~↑}}{} & \GACScoreCell{\GACMetricHeader{PDMS}{PDMS~↑}}{\GACMetricHeader{PDMS}{PDMS~↑}}{} & \GACScoreCell{\GACMetricHeader{RC}{RC~↑}}{\GACMetricHeader{RC}{RC~↑}}{} & \GACScoreCell{\GACMetricHeader[72pt]{HD-Score}{HD-Score~↑}}{\GACMetricHeader[72pt]{HD-Score}{HD-Score~↑}}{}\\
\midrule
\GACMethodCitation{iDriveVLA}{he2026idrivevla} & \CLIDriveOriginal\\
\GACRowColor
\hspace{0.8em}+ GAC & \CLIDriveGACWithChanges\\
\midrule
\GACMethodCitation{DrivoR}{kirby2026drivor} & \CLDrivoROriginal\\
\GACRowColor
\hspace{0.8em}+ GAC & \CLDrivoRGACWithChanges\\
\midrule
\GACMethodCitation{DiffusionDrive}{liao2025diffusiondrive} & \CLDiffusionOriginal\\
\GACRowColor
\hspace{0.8em}+ GAC & \CLDiffusionGACWithChanges\\
\bottomrule
\GACEndTabularStar
\end{table*}

%% file: figures/Figure4_layout.tex
\begin{figure*}[t]
\centering
\includegraphics[width=\linewidth]{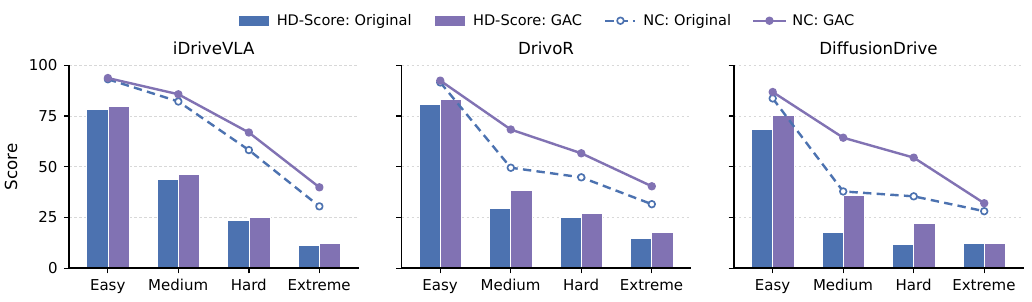}
\caption{Closed-loop performance across four difficulty levels on HUGSIM-160, with 40 scenarios per level drawn from four data sources. Bars show HD-Score and lines show NC, with Original and GAC shown in blue and purple, respectively.}
\label{fig:closed_loop_difficulty}
\end{figure*}

%% file: tables/Table3.tex
\begin{table}[t]
\centering
\caption{Effect of assessment frequency on HUGSIM-64. GAC performs assessment at 1 or 2\,Hz, while all planners operate at 4\,Hz. Purple values indicate changes relative to Original.}
\label{tab:frequency}
\GACTableStyle
\setlength{\tabcolsep}{3pt}
\GACBeginTabularStar{\linewidth}{@{}>{\raggedright\arraybackslash}m{82pt}@{\extracolsep{\fill}}*{3}{l}@{\hspace{10pt}}}{@{}l*{3}{c}@{}}
\toprule
\textbf{Method} & \GACFrequencyCell{\GACMetricHeader{PDMS}{PDMS~↑}}{} & \GACFrequencyCell{\GACMetricHeader{RC}{RC~↑}}{} & \GACFrequencyCell{\GACMetricHeader[72pt]{HD-Score}{HD-Score~↑}}{}\\
\midrule
\GACMethodCitation{iDriveVLA}{he2026idrivevla} & \FreqIDriveOriginal\\
\hspace{0.8em}+ GAC (1\,Hz) & \FreqIDriveOneHzWithChanges\\
\GACRowColor
\hspace{0.8em}+ GAC (2\,Hz) & \FreqIDriveTwoHzWithChanges\\
\midrule
\GACMethodCitation{DrivoR}{kirby2026drivor} & \FreqDrivoROriginal\\
\hspace{0.8em}+ GAC (1\,Hz) & \FreqDrivoROneHzWithChanges\\
\GACRowColor
\hspace{0.8em}+ GAC (2\,Hz) & \FreqDrivoRTwoHzWithChanges\\
\midrule
\GACMethodCitation{DiffusionDrive}{liao2025diffusiondrive} & \FreqDiffusionOriginal\\
\hspace{0.8em}+ GAC (1\,Hz) & \FreqDiffusionOneHzWithChanges\\
\GACRowColor
\hspace{0.8em}+ GAC (2\,Hz) & \FreqDiffusionTwoHzWithChanges\\
\bottomrule
\GACEndTabularStar
\par\vspace{-9pt}% Local adjustment of the table-to-text gap.
\end{table}

%% file: tables/Table4.tex
\begin{table*}[t]
\centering
\caption{Open-loop performance on NAVSIM v1. Failed includes all eligible zero-score scenes, while Intermediate and High-quality results are averaged across three sampled sets per planner. Purple values show PDMS changes from Original.}
\label{tab:combined_components}
\GACTableStyle
\setlength{\tabcolsep}{0.8pt}
\GACBeginTabularStar{\textwidth}{@{}>{\raggedright\arraybackslash}m{66pt}@{\hspace{4pt}\extracolsep{\fill}}*{3}{c}l@{\hspace{4pt}}*{3}{c}l@{}*{3}{c}l@{}}{@{}l*{12}{c}@{}}
\toprule
\raisebox{-0.75\baselineskip}[0pt][0pt]{\textbf{Method}} & \multicolumn{4}{c}{\textbf{Failed}\hspace*{6pt}} & \multicolumn{4}{c}{\textbf{Intermediate}\hspace*{6pt}} & \multicolumn{4}{c}{\textbf{High-quality}}\\
\cmidrule(l{0pt}r{6pt}){2-5}\cmidrule(l{0pt}r{6pt}){6-9}\cmidrule(l{0pt}r{0pt}){10-13}
& \GACAtomicHeader{\GACMetricHeader{NC}{NC~↑}} & \GACAtomicHeader{\GACMetricHeader{DAC}{DAC~↑}} & \GACAtomicHeader{\GACMetricHeader{TTC}{TTC~↑}} & \GACAtomicHeader{\GACMetricHeader{PDMS}{PDMS~↑}} & \GACAtomicHeader{\GACMetricHeader{NC}{NC~↑}} & \GACAtomicHeader{\GACMetricHeader{DAC}{DAC~↑}} & \GACAtomicHeader{\GACMetricHeader{TTC}{TTC~↑}} & \GACAtomicHeader{\GACMetricHeader{PDMS}{PDMS~↑}} & \GACAtomicHeader{\GACMetricHeader{NC}{NC~↑}} & \GACAtomicHeader{\GACMetricHeader{DAC}{DAC~↑}} & \GACAtomicHeader{\GACMetricHeader{TTC}{TTC~↑}} & \GACAtomicHeader{\GACMetricHeader{PDMS}{PDMS~↑}}\\
\midrule
\GACMethodCitation{iDriveVLA}{he2026idrivevla} & 33.51 & 65.96 & 34.57 & \GACScoreCell{\GACMetricHeader{PDMS}{PDMS~↑}}{0.00}{} & 99.33 & 100.00 & 75.00 & \GACScoreCell{\GACMetricHeader{PDMS}{PDMS~↑}}{78.02}{} & 100.00 & 100.00 & 100.00 & \GACScoreCell[20pt]{\GACMetricHeader{PDMS}{PDMS~↑}}{98.41}{}\\
\GACRowColor
\hspace{0.8em}+ GAC & 72.34 & 79.26 & 68.09 & \GACScoreCell{\GACMetricHeader{PDMS}{PDMS~↑}}{47.41}{+47.41} & 99.67 & 99.33 & 87.00 & \GACScoreCell{\GACMetricHeader{PDMS}{PDMS~↑}}{81.56}{+3.54} & 99.67 & 100.00 & 99.00 & \GACScoreCell[20pt]{\GACMetricHeader{PDMS}{PDMS~↑}}{96.22}{-2.19}\\
\midrule
\GACMethodCitation{DrivoR}{kirby2026drivor} & 50.23 & 48.40 & 47.49 & \GACScoreCell{\GACMetricHeader{PDMS}{PDMS~↑}}{0.00}{} & 99.50 & 100.00 & 85.33 & \GACScoreCell{\GACMetricHeader{PDMS}{PDMS~↑}}{80.82}{} & 100.00 & 100.00 & 100.00 & \GACScoreCell[20pt]{\GACMetricHeader{PDMS}{PDMS~↑}}{97.94}{}\\
\GACRowColor
\hspace{0.8em}+ GAC & 73.52 & 69.86 & 69.86 & \GACScoreCell{\GACMetricHeader{PDMS}{PDMS~↑}}{40.36}{+40.36} & 99.83 & 98.67 & 90.33 & \GACScoreCell{\GACMetricHeader{PDMS}{PDMS~↑}}{81.22}{+0.40} & 99.67 & 99.00 & 99.67 & \GACScoreCell[20pt]{\GACMetricHeader{PDMS}{PDMS~↑}}{95.10}{-2.84}\\
\midrule
\GACMethodCitation{DiffusionDrive}{liao2025diffusiondrive} & 67.28 & 31.24 & 63.40 & \GACScoreCell{\GACMetricHeader{PDMS}{PDMS~↑}}{0.00}{} & 99.67 & 100.00 & 83.67 & \GACScoreCell{\GACMetricHeader{PDMS}{PDMS~↑}}{81.31}{} & 100.00 & 100.00 & 100.00 & \GACScoreCell[20pt]{\GACMetricHeader{PDMS}{PDMS~↑}}{96.60}{}\\
\GACRowColor
\hspace{0.8em}+ GAC & 85.21 & 56.19 & 75.60 & \GACScoreCell{\GACMetricHeader{PDMS}{PDMS~↑}}{36.26}{+36.26} & 99.67 & 98.67 & 89.67 & \GACScoreCell{\GACMetricHeader{PDMS}{PDMS~↑}}{81.94}{+0.63} & 100.00 & 99.67 & 99.33 & \GACScoreCell[20pt]{\GACMetricHeader{PDMS}{PDMS~↑}}{95.14}{-1.46}\\
\bottomrule
\GACEndTabularStar
\par\vspace{-3pt}% Local adjustment of the table-to-text gap.
\end{table*}

%% file: figures/Figure5_layout.tex
\begin{figure*}[t]
\centering
\iflatexml\else
\begingroup%
\phantomsubcaption\label{fig:qualitative_open}%
\phantomsubcaption\label{fig:qualitative_closed}%
\endgroup%
\fi%
\iflatexml
\includegraphics[width=\linewidth,alt={(a) Open-loop recovery on NAVSIM v1. (b) Closed-loop driving on HUGSIM.}]{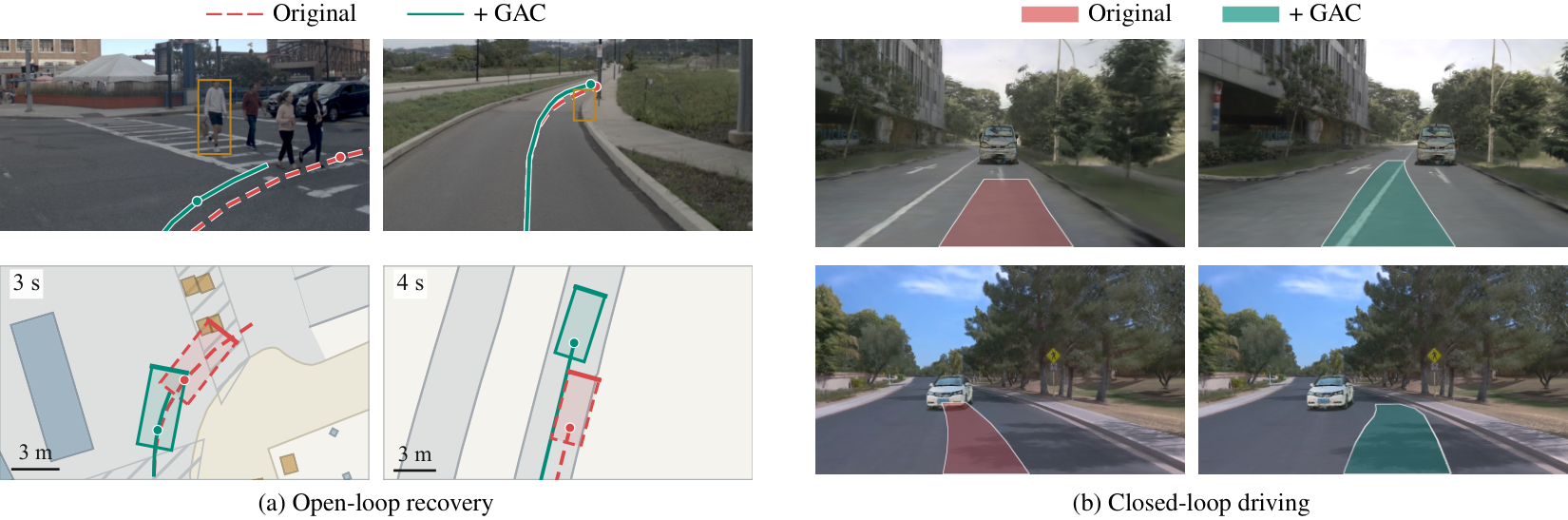}%
\else
\includegraphics[width=\linewidth,alt={(a) Open-loop recovery on NAVSIM v1. (b) Closed-loop driving on HUGSIM.}]{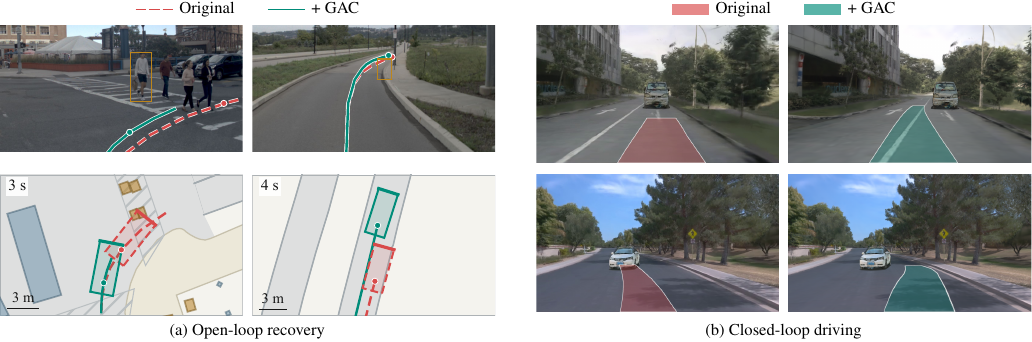}%
\fi%
\caption{Qualitative comparisons on NAVSIM v1 and HUGSIM. (a) Open-loop recovery at a pedestrian crossing (left) and a narrow bend (right). Camera views (top) and bird's-eye views (BEVs, bottom) visualize the trajectories selected by Original and GAC. Markers and footprints indicate $t=3$\,s on the left and $t=4$\,s on the right. BEVs are used only for analysis and are not provided to the critic. (b) Closed-loop comparisons at nearby ego positions, with ribbons showing projected full vehicle footprints. Red and green denote Original and GAC, respectively. All trajectories span four seconds.}
\label{fig:qualitative}
\end{figure*}

%% file: tables/Table5.tex
\begin{table}[t]
\centering
\caption{Workflow and input ablations on matched iDriveVLA sets. (a) Uniform samples a candidate uniformly after rejection, with results reported as exact expectations. Post reviews the selected candidate against the original selection. Bold and underlined values indicate the best and second-best results per column within each subtable, respectively.}
\label{tab:harness_ablations}
\GACTableStyle
\setlength{\tabcolsep}{1.0pt}
\GACBeginTabular{@{}>{\centering\arraybackslash}m{45pt}>{\centering\arraybackslash}m{45pt}>{\centering\arraybackslash}m{40pt}>{\centering\arraybackslash}m{28pt}>{\centering\arraybackslash}m{40pt}>{\centering\arraybackslash}m{28pt}@{}}{@{}*{6}{c}@{}}
\toprule
\multicolumn{2}{@{}l}{\textbf{(a) Workflow ablations}} & \multicolumn{2}{c}{\textbf{Failed}} & \multicolumn{2}{c}{\textbf{High-quality}}\\
\cmidrule(l{0pt}r{1pt}){1-2}\cmidrule(l{1pt}r{1pt}){3-4}\cmidrule(l{1pt}r{0pt}){5-6}
\textbf{Assess} & \textbf{Select} & \multicolumn{1}{c}{\GACFixedHeader{40pt}{\GACMetricHeader{NC$\times$DAC}{NC×DAC~↑}}} & \multicolumn{1}{c}{\GACFixedHeader{28pt}{\GACMetricHeader{PDMS}{PDMS~↑}}} & \multicolumn{1}{c}{\GACFixedHeader{40pt}{\GACMetricHeader{NC$\times$DAC}{NC×DAC~↑}}} & \multicolumn{1}{c@{}}{\GACFixedHeader{28pt}{\GACMetricHeader{PDMS}{PDMS~↑}}}\\
\midrule
$\checkmark$ & Uniform & 25.70 & 22.55 & \underline{98.78} & \underline{95.90}\\
 & $\checkmark$ & \textbf{65.59} & \textbf{61.15} & 91.00 & 87.94\\
Post & $\checkmark$ & 48.39 & 43.46 & 93.00 & 90.65\\
\GACRowColor
$\checkmark$ & $\checkmark$ & \underline{53.76} & \underline{49.26} & \textbf{99.00} & \textbf{95.95}\\
\midrule
\multicolumn{2}{@{}l}{\textbf{(b) Input ablations}} & \multicolumn{2}{c}{\textbf{Failed}} & \multicolumn{2}{c}{\textbf{High-quality}}\\
\cmidrule(l{0pt}r{1pt}){1-2}\cmidrule(l{1pt}r{1pt}){3-4}\cmidrule(l{1pt}r{0pt}){5-6}
\textbf{Overlays} & \textbf{Principles} & \multicolumn{1}{c}{\GACFixedHeader{40pt}{\GACMetricHeader{NC$\times$DAC}{NC×DAC~↑}}} & \multicolumn{1}{c}{\GACFixedHeader{28pt}{\GACMetricHeader{PDMS}{PDMS~↑}}} & \multicolumn{1}{c}{\GACFixedHeader{40pt}{\GACMetricHeader{NC$\times$DAC}{NC×DAC~↑}}} & \multicolumn{1}{c@{}}{\GACFixedHeader{28pt}{\GACMetricHeader{PDMS}{PDMS~↑}}}\\
\midrule
$\checkmark$ & & \underline{32.26} & \underline{29.22} & \textbf{99.00} & \textbf{96.87}\\
 & $\checkmark$ & 30.11 & 27.42 & \underline{97.00} & 93.82\\
\GACRowColor
$\checkmark$ & $\checkmark$ & \textbf{53.76} & \textbf{49.26} & \textbf{99.00} & \underline{95.95}\\
\bottomrule
\end{tabular}
\par\vspace{-7pt}% Local adjustment of the table-to-text gap.
\end{table}

%% file: sections/conclusion.tex
\section{Conclusion}
We introduce Generalist Action Critic (GAC), a new paradigm for zero-shot, inference-time cooperation between frontier models and specialized driving planners. Through an evidence-grounded inference harness, GAC integrates layered visual representations and a structured reasoning context to support selective decision revision. Closed-loop experiments on HUGSIM demonstrate consistent improvements in driving safety and overall performance across three planners. GAC also enables DrivoR to achieve the highest HD-Score among the compared methods on the full HUGSIM benchmark. On NAVSIM v1, GAC recovers up to 52.13\% of eligible zero-score failures. Ablation studies further validate the effectiveness of GAC's key components. Together, these findings demonstrate that frontier-model reasoning can complement specialized driving planners through inference-time cooperation to improve driving safety.

%% file: sections/acknowledgments.tex
\section*{Acknowledgments}
This study is supported by the National Natural Science Foundation of China (Grant Nos. 92582205 and 52602738), the Shanghai Automotive Industry Science and Technology Development Foundation, and the Tsinghua University (State Key Laboratory of Intelligent Green Vehicle and Mobility)--Xiaomi EV Technology Co., Ltd Joint Research Center for Autonomous Driving Safety and Intelligence.

%% file: sections/appendix.tex
\section{Implementation Details}\label{app:implementation}

\subsection{Geometric Projection and Rendering}\label{app:geometry_details}\label{app:lens_mapping}
At $t=0$, the ego coordinate frame is centered at the trajectory reference point, with the $x$-axis pointing forward and the $y$-axis pointing left. Candidate trajectories, spatial anchors, and future occupancy are all expressed in this fixed frame. We assume a flat ground plane at $z=0$.

\paragraph{Spatial anchors.}
We construct an axis-aligned grid extending from $-40$ to $40$\,m along both ego axes, with grid lines spaced $5$\,m apart. The resulting grid contains $16\times16$ cells, each measuring $5\,\mathrm m\times5\,\mathrm m$.

\paragraph{Vehicle footprint union.}
For the planner's originally selected trajectory $\tau_b$, we construct the vehicle's occupied region over the prediction horizon. The vehicle's ground footprint is represented as a filled rectangle defined relative to the trajectory reference point:
\begin{equation}
B=[-\ell_r,\ell_f]\times[-w/2,w/2]\subset\mathbb R^2,
\label{eq:vehicle_footprint}
\end{equation}
where $\ell_r$ and $\ell_f$ are the distances from the reference point to the rear and front edges of the vehicle, respectively, and $w$ is the vehicle width.

At time $t$, the candidate's reference-point position $\mathbf q_b^t=(x_b^t,y_b^t)^\top$ and heading $\psi_b^t$ define the transformation of a vehicle-local point $\mathbf p=(x,y)^\top\in B$ into the fixed ego frame:
\begin{equation}
\phi^t(\mathbf p)=\mathbf R(\psi_b^t)\mathbf p+\mathbf q_b^t,
\label{eq:footprint_transform}
\end{equation}
where $\mathbf R(\psi)$ is the two-dimensional rotation matrix for heading $\psi$. The original trajectory consists of eight future poses at $0.5$-second intervals up to $4$ seconds, together with the initial pose $(0,0,0)$ at $t=0$. We linearly interpolate positions and unwrapped headings at the sampling times $\mathcal T=\{0,0.02,\ldots,4.0\}\,\mathrm{s}$. The footprint union is defined as
\begin{equation}
\mathcal U=\bigcup_{t\in\mathcal T}\{\phi^t(\mathbf p):\mathbf p\in B\}.
\label{eq:sampled_footprint_union}
\end{equation}
The resulting occupancy layer is rendered in pink and corresponds to the originally selected trajectory in both critic stages.

\paragraph{Camera projection.}
Let $v\in\{\mathrm F,\mathrm L,\mathrm R\}$ denote the front, front-left, and front-right cameras. We project the spatial anchors and vehicle footprint union into each camera view using the calibrated ego-to-camera rotation $\mathbf R_v$ and translation $\mathbf t_v$. Projection follows a zero-skew pinhole model with intrinsic matrix $\mathbf K_v$, focal lengths $f_{x,v},f_{y,v}$, and principal point $(c_{x,v},c_{y,v})$. The intrinsics are scaled to the resolution of the raw camera images. Only points with positive camera depth $Z_v>0$ are projected, yielding ideal pixel coordinates $(u_x,u_y)$.

\paragraph{Theoretical horizon.}
The theoretical horizon is the vanishing line of the ground plane, corresponding to viewing rays parallel to the ground. Let $\mathbf n_v=(n_{x,v},n_{y,v},n_{z,v})^\top=\mathbf R_v(0,0,1)^\top$ denote the ground normal in camera coordinates. For $n_{y,v}\neq0$, the vertical coordinate of the horizon at horizontal pixel coordinate $u_x$ is
\begin{equation}
u_y=c_{y,v}-\frac{f_{y,v}}{n_{y,v}}\left[\frac{n_{x,v}}{f_{x,v}}(u_x-c_{x,v})+n_{z,v}\right].
\label{eq:ground_horizon}
\end{equation}
The resulting horizon is a straight line in the ideal image. We render its visible portion as a red dashed line on the annotation canvas.

\paragraph{Rendering and compositing.}
Anchor segments and vehicle footprints are first clipped against a near plane. The resulting geometry is then projected into the ideal image and rendered alongside the theoretical horizon on annotation layers. To match the lens distortion of the raw camera images, we use a calibrated five-parameter Brown--Conrady model~\citep{brown1966decentering} where required. The annotation canvases are enlarged beyond the image boundaries to accommodate the displacement of projected geometry caused by distortion. The shifted canvas origins are accounted for in the principal points and inverse-sampling coordinates. We then use inverse mapping to resample the annotations into the raw image coordinates and composite them onto the original camera images without undistorting or geometrically resampling those images.

% Declare the request figure early so it can appear beside the A.2 discussion.
\begin{figure*}[tp]
\centering
\iflatexml
\GACHTMLFullWidthGraphic{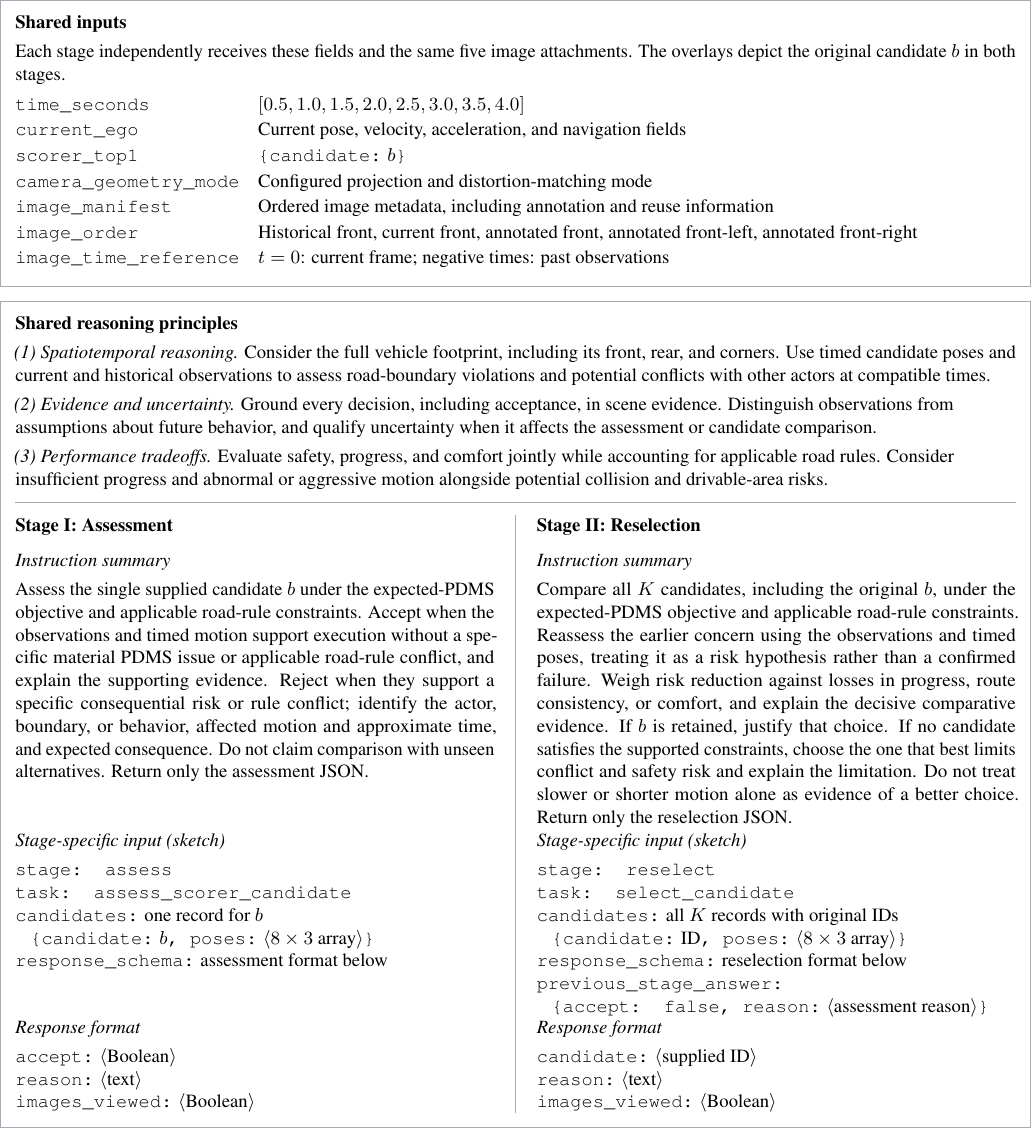}
\else
\includegraphics[width=\textwidth]{figures/Figure6.pdf}
\fi
\caption{Critic context and response formats for GAC on NAVSIM v1. Each stage independently receives the shared inputs and reasoning principles, together with stage-specific instructions and candidate inputs. Instructions are summarized, and angle brackets denote placeholders.}
\label{fig:critic_request_structure}
\end{figure*}

\subsection{Critic Context and Response Formats}\label{app:textual_input}\label{app:context_formats}
Each critic stage runs in a fresh Codex context using GPT-6 Astra~\citep{openai2026astra} with maximum reasoning effort. Each call receives complete stage instructions, a JSON-serialized scene payload, image attachments, and the corresponding response schema, without inheriting conversation or image memory from the other stage. Figure~\ref{fig:critic_request_structure} summarizes the request structure and response formats. The same structure is used for HUGSIM, with adaptations to its coordinate conventions, evaluation metrics, and camera geometry.

\paragraph{Images and metadata.}
The five attachments consist of a historical front image captured $1.5$ seconds earlier, the current front image, and annotated current front, front-left, and front-right images. Both stages receive the same images, with the annotated views combining spatial anchors $A$ and the original candidate's occupancy $O$. The accompanying metadata specify the viewpoints, observation times, geometric annotations, candidate identifiers, and camera projection conventions.

\paragraph{Candidate inputs and selection.}
Candidate trajectories are supplied as time-indexed $[x,y,\psi]$ poses using the coordinate conventions of each benchmark. Assessment receives only the originally selected trajectory $\tau_b$, while reselection receives the complete candidate set $\mathcal C$ together with the previous assessment decision and reason. The candidate count is $64$ for iDriveVLA and DrivoR and $20$ for DiffusionDrive. The original selection is determined by the planner's candidate scores, with ties resolved by candidate index. The critic does not receive the planners' full ranking-score vectors, official evaluation labels, or ground-truth information.

\paragraph{Response validation and execution.}
Each valid response contains exactly the fields shown in Figure~\ref{fig:critic_request_structure}. Assessment requires a Boolean acceptance decision, while reselection requires a candidate identifier from the supplied set. Both stages require a nonempty reason and a self-reported confirmation of image inspection. The reselection reason explains how the earlier concern is addressed and, when the original candidate $b$ is retained, justifies that choice. A validated acceptance returns $b$ immediately, while a validated rejection triggers reselection, whose selected candidate determines the final choice.

\section{Experimental Details}\label{app:experimental_details}

\subsection{Checkpoints and Benchmark Adaptations}\label{app:checkpoints_adaptations}
We use an unreleased iDriveVLA checkpoint~\citep{he2026idrivevla}, the publicly released DrivoR NAVSIM v1 checkpoint without SimScale~\citep{kirby2026drivor}, and the publicly released DiffusionDrive NAVSIM checkpoint~\citep{liao2025diffusiondrive}. All three planners use front, front-left, and front-right images together with ego state. iDriveVLA and DrivoR additionally use rear-view images, while DiffusionDrive uses LiDAR. For candidate selection, iDriveVLA and DrivoR use learned planning-quality scores, and DiffusionDrive uses classification scores. The critic uses a consistent five-image configuration across planners, together with stage-specific candidate information. For NAVSIM, candidate trajectories are exported before critic evaluation and reused across selection methods. DiffusionDrive uses fixed per-scene randomness during candidate generation.

\paragraph{Sensor and geometry adaptation.}
For NAVSIM, vehicle footprints use the nuPlan Pacifica dimensions in the rear-axle frame, with $\ell_f=4.049\,\mathrm m$, $\ell_r=1.127\,\mathrm m$, and $w=2.297\,\mathrm m$. For HUGSIM, we use the ground projection of the vehicle center as the reference point for candidate poses, spatial anchors, and vehicle footprints, with camera calibration defined consistently in the same frame. The footprint dimensions are obtained from scene metadata, with $\ell_f=\ell_r=L/2$, where $L$ and $w$ denote vehicle length and width. HUGSIM uses a pinhole camera model for image rendering, allowing geometric annotations to be projected directly without lens-distortion matching. For DiffusionDrive on HUGSIM, we generate a synthetic point cloud from rendered depth, using the same adaptation for Original and GAC.

\subsection{Scenario Selection and Evaluation Protocols}\label{app:evaluation_protocols}\label{app:closed_loop_protocol}
\paragraph{HUGSIM scenarios and execution.}
For the full benchmark comparison, we evaluate DrivoR + GAC with assessment at 2\,Hz on the 345-scenario pre-challenge HUGSIM~\citep{zhou2024hugsim} evaluation set used by DrivoR~\citep{kirby2026drivor}. The set contains 48 Easy, 127 Medium, 86 Hard, and 84 Extreme scenarios. We compare GAC with previously published results from DriveZero~\citep{xiaomi2026drivezero} and DrivoR~\citep{kirby2026drivor} under the same evaluation protocol, with overall scores averaged equally across all 345 scenarios.

For cross-planner evaluation, we construct HUGSIM-160 by randomly sampling 10 scenarios from each combination of four data sources (KITTI-360, nuScenes, PandaSet, and Waymo) and four difficulty levels (Easy, Medium, Hard, and Extreme). All three planners are evaluated on the same scenarios, with GAC and Original compared under identical settings. GAC assessment runs at 2\,Hz on HUGSIM-160. We further construct HUGSIM-64 by randomly sampling four scenarios from each data source and difficulty combination within HUGSIM-160. This subset is evaluated at both 1\,Hz and 2\,Hz to compare critic assessment frequencies.

The simulator and planner operate at 4\,Hz, with GAC reviewing the planner's selection every two planning steps at 2\,Hz or every four steps at 1\,Hz. At each planning update, the planner generates new candidates from the current observations, and only the initial segment of the selected trajectory is executed. The planner's original selection is used at steps without critic assessment, and the simulator waits for the critic response when assessment is triggered. All frequencies refer to simulation time, with simulator and planner random seeds fixed at zero. Missing historical observations are explicitly indicated in the image attachments.

\paragraph{NAVSIM scenarios.}
The Failed group includes original selections with zero PDMS and at least one positive-score alternative. All 188, 219, and 541 eligible scenes are included for iDriveVLA, DrivoR, and DiffusionDrive, respectively. The Intermediate and High-quality groups are defined by original PDMS scores in $(0,90)$ and $[90,100]$, respectively. For each planner, we independently sample three sets of 100 distinct scenes from each of the Intermediate and High-quality groups. Scenes may appear in multiple sets, and each occurrence is included in the reported statistics.

\subsection{Scoring Formulas and Aggregation}\label{app:scoring_aggregation}\label{app:closed_loop_metrics}
\paragraph{HUGSIM.}
HUGSIM~\citep{zhou2024hugsim} evaluates no collision (NC), drivable area compliance (DAC), time to collision within bound (TTC), comfort (C), and route completion (RC). On the native 0--1 scale, the score at evaluated planning step $t$ in scenario $s$ is
\begin{equation}
q_{s,t}=\mathrm{NC}_{s,t}\cdot\mathrm{DAC}_{s,t}\,
\frac{5\,\mathrm{TTC}_{s,t}+2\,\mathrm{C}_{s,t}}{7}.
\end{equation}
The episode PDMS and HD-Score are
\begin{equation}
\bar q_s=\frac{1}{T_s}\sum_{t=1}^{T_s}q_{s,t},
\qquad h_s=\bar q_s\cdot\mathrm{RC}_s,
\end{equation}
where $T_s$ is the number of evaluated planning steps and $\mathrm{RC}_s$ measures route completion over the episode. NC, DAC, TTC, and C are averaged over evaluated planning steps within each episode and then equally across episodes, while episode PDMS, RC, and HD-Score are averaged equally across scenarios. All reported scores are scaled to 0--100.

\paragraph{NAVSIM.}
We use the official NAVSIM v1 PDMS~\citep{dauner2024navsim}, which combines no at-fault collision (NC), drivable area compliance (DAC), ego progress (EP), time to collision (TTC), and comfort (C). On the reported 0--100 scale, the score is
\PDMSDefinition
PDMS is computed for each evaluated instance before averaging. The reported $\mathrm{NC}\times\mathrm{DAC}$ score is obtained by multiplying the two component scores within each instance before averaging and scaling to 0--100. TTC and EP are averaged over all evaluated instances, including those with zero final PDMS.

\section{Additional Closed-Loop Results}\label{app:closed_loop}

\subsection{Detailed Results by Data Source and Difficulty}\label{app:closed_loop_breakdown}
Tables~\ref{tab:closed_loop_datasets} and~\ref{tab:closed_loop_difficulties} report results on HUGSIM-160 by data source and difficulty, with 40 scenarios per group. Across all three planners, GAC consistently improves NC and TTC in every data source and difficulty group, alongside gains in mean PDMS and HD-Score. Medium scenarios yield the largest absolute HD-Score gains for all three planners. DAC improvements are less consistent, which may reflect the difficulty of precise drivable-area boundary assessment. In some Hard and Extreme groups, higher PDMS is accompanied by modest reductions in route completion, while HD-Score still improves.

\input{tables/Table6}

\input{tables/Table7}

\subsection{Component Scores by Assessment Frequency}\label{app:closed_loop_frequency}
Table~\ref{tab:closed_loop_frequency_components} reports NC, DAC, TTC, and comfort on HUGSIM-64, complementing the PDMS, RC, and HD-Score results in Table~\ref{tab:frequency}. Increasing the critic assessment frequency from 1\,Hz to 2\,Hz improves NC and TTC across all three planners, alongside the gains in PDMS and HD-Score.

\input{tables/Table8}

\section{Open-Loop Analysis}\label{app:additional_results}\label{app:experiment_records}

\subsection{Detailed Results and Sampling Variability}\label{app:open_loop_detailed}
Table~\ref{tab:sampled_variability} reports mean scores and twice the sample standard deviation of the three set-level means for the Intermediate and High-quality groups. Among the 300 High-quality evaluation entries per planner, GAC introduces one new zero-score failure for iDriveVLA, four for DrivoR, and one for DiffusionDrive. Comfort scores are 100 for both Original and GAC across all planners and groups and are therefore omitted from the table.

\input{tables/Table9}
\subsection{Assessment and Reselection Analysis}\label{app:assessment_reselection}
\begin{figure}[t]
\centering
\includegraphics[width=\linewidth]{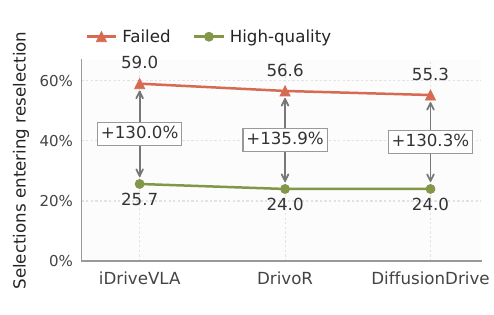}
\caption{Stage I rejection rates for Failed and High-quality selections on NAVSIM v1. Point labels show the percentages passed to Stage II. Arrows indicate the relative increase in rejection rate from High-quality to Failed for each planner.}
\label{fig:intervention_analysis}\label{fig:reselection_entry}
\end{figure}

Figure~\ref{fig:intervention_analysis} summarizes Stage I rejection rates for the Failed and High-quality groups. Assessment triggers reselection for 55.27--59.04\% of Failed selections, compared with 24.00--25.67\% of High-quality selections, representing a relative increase of up to 136\% over the High-quality group. Among the Failed selections rejected in Stage I, GAC recovers 88.29\%, 78.23\%, and 76.92\% for iDriveVLA, DrivoR, and DiffusionDrive, respectively, demonstrating strong recovery performance on identified failures.

\input{sections/supplementary_temporal_analysis}

\section{Ablation Details}\label{app:ablation_configurations}

\subsection{Workflow Ablations}\label{app:workflow_ablations}
\emph{Original} uses the planner's original selection without critic intervention. \emph{Direct} bypasses assessment and selects from all 64 candidates in a single call. It retains the reasoning principles and five-image configuration, including spatial anchors but excluding the original candidate's footprint overlays, explicit original-selection information, and assessment feedback. \emph{Select then review} first selects a candidate independently and then compares it with the planner's original selection to determine the final choice between the two. This variant preserves the reasoning principles, spatial anchors, and original candidate's footprint overlays.

\emph{Uniform} retains GAC's recorded assessment decisions and accepted selections, replacing only the reselection strategy after rejection. It selects from all 64 candidates with equal probability, including the original candidate, without additional reselection model calls.

We compute the expected PDMS of \emph{Uniform} analytically. For each scene $s$, let $a(s)=1$ indicate acceptance and $a(s)=0$ rejection, with $q_b(s)$ denoting the original selection's PDMS and $q_i(s)$ the score of candidate $i$ among $K_s$ candidates. The expected final score is
\begin{equation}
\mathbb E[q_{\mathrm U}(s)]=a(s)q_b(s)+\bigl(1-a(s)\bigr)\frac{1}{K_s}\sum_{i=0}^{K_s-1}q_i(s).
\end{equation}
Averaging these expectations across the evaluated scenes gives the exact expected mean PDMS.

\subsection{Input Ablations}\label{app:input_ablations}
The ablation without reasoning principles retains the field definitions, coordinate and image conventions, evaluation definitions, and response schemas. The ablation without geometric overlays removes spatial anchors and projected vehicle footprints from the images while retaining the reasoning principles. Both configurations preserve GAC's assessment and conditional reselection procedure.

\section{Qualitative Analysis}\label{app:qualitative}
BEVs, candidate-set plots, and evaluator scores are used only for analysis and are not supplied to the critic.

\subsection{Visual Evidence and Critic Reasoning}\label{app:context_effects}
\input{sections/reasoning_cases}

\subsection{Closed-Loop Driving}\label{app:closed_loop_additional_case}
\input{sections/closed_loop_cases}

\subsection{Open-Loop Recovery}\label{app:open_loop_cases}
\input{sections/open_loop_cases}

%% file: tables/Table6.tex
\begin{table*}[tp]
\centering
\caption{Closed-loop performance by data source on HUGSIM-160. Each source contains 40 scenarios. Shading marks GAC; purple values show score changes from Original.}
\label{tab:closed_loop_datasets}
\GACTableStyle
\setlength{\tabcolsep}{2pt}
\GACBeginTabularStar{\linewidth}{@{}>{\raggedright\arraybackslash}m{43pt}>{\raggedright\arraybackslash}m{56pt}@{\extracolsep{\fill}}*{7}{l}@{}}{@{}ll*{7}{c}@{}}
\toprule
\textbf{Data source} & \textbf{Method} & \GACFixedCell{22pt}{\GACMetricHeader{NC}{NC~↑}}{28pt}{} & \GACFixedCell{22pt}{\GACMetricHeader{DAC}{DAC~↑}}{28pt}{} & \GACFixedCell{22pt}{\GACMetricHeader{TTC}{TTC~↑}}{28pt}{} & \GACFixedCell{22pt}{\GACMetricHeader{Comfort}{Comfort~↑}}{28pt}{} & \GACFixedCell{22pt}{\GACMetricHeader{PDMS}{PDMS~↑}}{28pt}{} & \GACFixedCell{22pt}{\GACMetricHeader{RC}{RC~↑}}{28pt}{} & \GACFixedCell{22pt}{\GACMetricHeader[72pt]{HD-Score}{HD-Score~↑}}{28pt}{}\\
\midrule
KITTI-360 & iDriveVLA & \GACFixedCell{22pt}{57.63}{28pt}{} & \GACFixedCell{22pt}{84.96}{28pt}{} & \GACFixedCell{22pt}{53.80}{28pt}{} & \GACFixedCell{22pt}{97.58}{28pt}{} & \GACFixedCell{22pt}{46.75}{28pt}{} & \GACFixedCell{22pt}{30.53}{28pt}{} & \GACFixedCell{22pt}{23.87}{28pt}{}\\
\GACPDFRowFill
 & \GACHTMLRowFill\hspace{0.8em}+ GAC & \GACFixedCell{22pt}{61.92}{28pt}{+4.29} & \GACFixedCell{22pt}{85.05}{28pt}{+0.09} & \GACFixedCell{22pt}{59.41}{28pt}{+5.61} & \GACFixedCell{22pt}{96.01}{28pt}{-1.57} & \GACFixedCell{22pt}{51.43}{28pt}{+4.68} & \GACFixedCell{22pt}{30.75}{28pt}{+0.22} & \GACFixedCell{22pt}{25.28}{28pt}{+1.41}\\
 & DrivoR & \GACFixedCell{22pt}{45.54}{28pt}{} & \GACFixedCell{22pt}{84.39}{28pt}{} & \GACFixedCell{22pt}{41.00}{28pt}{} & \GACFixedCell{22pt}{92.49}{28pt}{} & \GACFixedCell{22pt}{36.28}{28pt}{} & \GACFixedCell{22pt}{35.55}{28pt}{} & \GACFixedCell{22pt}{24.91}{28pt}{}\\
\GACPDFRowFill
 & \GACHTMLRowFill\hspace{0.8em}+ GAC & \GACFixedCell{22pt}{55.27}{28pt}{+9.73} & \GACFixedCell{22pt}{85.39}{28pt}{+1.00} & \GACFixedCell{22pt}{51.69}{28pt}{+10.69} & \GACFixedCell{22pt}{92.55}{28pt}{+0.06} & \GACFixedCell{22pt}{46.34}{28pt}{+10.06} & \GACFixedCell{22pt}{36.59}{28pt}{+1.04} & \GACFixedCell{22pt}{28.27}{28pt}{+3.36}\\
 & DiffusionDrive & \GACFixedCell{22pt}{30.35}{28pt}{} & \GACFixedCell{22pt}{79.90}{28pt}{} & \GACFixedCell{22pt}{24.61}{28pt}{} & \GACFixedCell{22pt}{96.80}{28pt}{} & \GACFixedCell{22pt}{18.44}{28pt}{} & \GACFixedCell{22pt}{17.42}{28pt}{} & \GACFixedCell{22pt}{6.63}{28pt}{}\\
\GACPDFRowFill
 & \GACHTMLRowFill\hspace{0.8em}+ GAC & \GACFixedCell{22pt}{41.29}{28pt}{+10.94} & \GACFixedCell{22pt}{84.56}{28pt}{+4.66} & \GACFixedCell{22pt}{34.13}{28pt}{+9.52} & \GACFixedCell{22pt}{92.78}{28pt}{-4.02} & \GACFixedCell{22pt}{29.27}{28pt}{+10.83} & \GACFixedCell{22pt}{22.41}{28pt}{+4.99} & \GACFixedCell{22pt}{12.98}{28pt}{+6.35}\\
\midrule
nuScenes & iDriveVLA & \GACFixedCell{22pt}{73.20}{28pt}{} & \GACFixedCell{22pt}{99.81}{28pt}{} & \GACFixedCell{22pt}{68.09}{28pt}{} & \GACFixedCell{22pt}{98.65}{28pt}{} & \GACFixedCell{22pt}{68.89}{28pt}{} & \GACFixedCell{22pt}{60.59}{28pt}{} & \GACFixedCell{22pt}{48.63}{28pt}{}\\
\GACPDFRowFill
 & \GACHTMLRowFill\hspace{0.8em}+ GAC & \GACFixedCell{22pt}{77.97}{28pt}{+4.77} & \GACFixedCell{22pt}{100.00}{28pt}{+0.19} & \GACFixedCell{22pt}{71.29}{28pt}{+3.20} & \GACFixedCell{22pt}{98.57}{28pt}{-0.08} & \GACFixedCell{22pt}{72.62}{28pt}{+3.73} & \GACFixedCell{22pt}{59.93}{28pt}{-0.66} & \GACFixedCell{22pt}{49.68}{28pt}{+1.05}\\
 & DrivoR & \GACFixedCell{22pt}{60.31}{28pt}{} & \GACFixedCell{22pt}{100.00}{28pt}{} & \GACFixedCell{22pt}{53.60}{28pt}{} & \GACFixedCell{22pt}{94.90}{28pt}{} & \GACFixedCell{22pt}{54.22}{28pt}{} & \GACFixedCell{22pt}{56.98}{28pt}{} & \GACFixedCell{22pt}{43.39}{28pt}{}\\
\GACPDFRowFill
 & \GACHTMLRowFill\hspace{0.8em}+ GAC & \GACFixedCell{22pt}{68.21}{28pt}{+7.90} & \GACFixedCell{22pt}{100.00}{28pt}{+0.00} & \GACFixedCell{22pt}{61.73}{28pt}{+8.13} & \GACFixedCell{22pt}{93.73}{28pt}{-1.17} & \GACFixedCell{22pt}{62.12}{28pt}{+7.90} & \GACFixedCell{22pt}{58.90}{28pt}{+1.92} & \GACFixedCell{22pt}{47.01}{28pt}{+3.62}\\
 & DiffusionDrive & \GACFixedCell{22pt}{60.70}{28pt}{} & \GACFixedCell{22pt}{98.02}{28pt}{} & \GACFixedCell{22pt}{54.13}{28pt}{} & \GACFixedCell{22pt}{89.90}{28pt}{} & \GACFixedCell{22pt}{52.71}{28pt}{} & \GACFixedCell{22pt}{53.02}{28pt}{} & \GACFixedCell{22pt}{37.77}{28pt}{}\\
\GACPDFRowFill
 & \GACHTMLRowFill\hspace{0.8em}+ GAC & \GACFixedCell{22pt}{72.74}{28pt}{+12.04} & \GACFixedCell{22pt}{97.82}{28pt}{-0.20} & \GACFixedCell{22pt}{63.40}{28pt}{+9.27} & \GACFixedCell{22pt}{91.54}{28pt}{+1.64} & \GACFixedCell{22pt}{63.18}{28pt}{+10.47} & \GACFixedCell{22pt}{62.33}{28pt}{+9.31} & \GACFixedCell{22pt}{48.76}{28pt}{+10.99}\\
\midrule
PandaSet & iDriveVLA & \GACFixedCell{22pt}{64.80}{28pt}{} & \GACFixedCell{22pt}{99.82}{28pt}{} & \GACFixedCell{22pt}{57.66}{28pt}{} & \GACFixedCell{22pt}{98.81}{28pt}{} & \GACFixedCell{22pt}{58.93}{28pt}{} & \GACFixedCell{22pt}{58.03}{28pt}{} & \GACFixedCell{22pt}{46.45}{28pt}{}\\
\GACPDFRowFill
 & \GACHTMLRowFill\hspace{0.8em}+ GAC & \GACFixedCell{22pt}{72.54}{28pt}{+7.74} & \GACFixedCell{22pt}{99.82}{28pt}{+0.00} & \GACFixedCell{22pt}{68.83}{28pt}{+11.17} & \GACFixedCell{22pt}{96.44}{28pt}{-2.37} & \GACFixedCell{22pt}{68.26}{28pt}{+9.33} & \GACFixedCell{22pt}{56.76}{28pt}{-1.27} & \GACFixedCell{22pt}{47.62}{28pt}{+1.17}\\
 & DrivoR & \GACFixedCell{22pt}{52.89}{28pt}{} & \GACFixedCell{22pt}{100.00}{28pt}{} & \GACFixedCell{22pt}{46.33}{28pt}{} & \GACFixedCell{22pt}{93.10}{28pt}{} & \GACFixedCell{22pt}{46.73}{28pt}{} & \GACFixedCell{22pt}{60.36}{28pt}{} & \GACFixedCell{22pt}{41.59}{28pt}{}\\
\GACPDFRowFill
 & \GACHTMLRowFill\hspace{0.8em}+ GAC & \GACFixedCell{22pt}{63.33}{28pt}{+10.44} & \GACFixedCell{22pt}{100.00}{28pt}{+0.00} & \GACFixedCell{22pt}{56.80}{28pt}{+10.47} & \GACFixedCell{22pt}{92.55}{28pt}{-0.55} & \GACFixedCell{22pt}{57.21}{28pt}{+10.48} & \GACFixedCell{22pt}{60.37}{28pt}{+0.01} & \GACFixedCell{22pt}{46.08}{28pt}{+4.49}\\
 & DiffusionDrive & \GACFixedCell{22pt}{50.76}{28pt}{} & \GACFixedCell{22pt}{99.80}{28pt}{} & \GACFixedCell{22pt}{45.18}{28pt}{} & \GACFixedCell{22pt}{85.17}{28pt}{} & \GACFixedCell{22pt}{44.68}{28pt}{} & \GACFixedCell{22pt}{57.47}{28pt}{} & \GACFixedCell{22pt}{38.96}{28pt}{}\\
\GACPDFRowFill
 & \GACHTMLRowFill\hspace{0.8em}+ GAC & \GACFixedCell{22pt}{64.22}{28pt}{+13.46} & \GACFixedCell{22pt}{99.96}{28pt}{+0.16} & \GACFixedCell{22pt}{57.83}{28pt}{+12.65} & \GACFixedCell{22pt}{86.74}{28pt}{+1.57} & \GACFixedCell{22pt}{56.62}{28pt}{+11.94} & \GACFixedCell{22pt}{59.44}{28pt}{+1.97} & \GACFixedCell{22pt}{46.71}{28pt}{+7.75}\\
\midrule
Waymo & iDriveVLA & \GACFixedCell{22pt}{68.45}{28pt}{} & \GACFixedCell{22pt}{96.85}{28pt}{} & \GACFixedCell{22pt}{63.89}{28pt}{} & \GACFixedCell{22pt}{98.01}{28pt}{} & \GACFixedCell{22pt}{62.24}{28pt}{} & \GACFixedCell{22pt}{46.30}{28pt}{} & \GACFixedCell{22pt}{38.14}{28pt}{}\\
\GACPDFRowFill
 & \GACHTMLRowFill\hspace{0.8em}+ GAC & \GACFixedCell{22pt}{73.86}{28pt}{+5.41} & \GACFixedCell{22pt}{97.01}{28pt}{+0.16} & \GACFixedCell{22pt}{69.61}{28pt}{+5.72} & \GACFixedCell{22pt}{96.60}{28pt}{-1.41} & \GACFixedCell{22pt}{68.48}{28pt}{+6.24} & \GACFixedCell{22pt}{48.23}{28pt}{+1.93} & \GACFixedCell{22pt}{41.09}{28pt}{+2.95}\\
 & DrivoR & \GACFixedCell{22pt}{58.72}{28pt}{} & \GACFixedCell{22pt}{99.44}{28pt}{} & \GACFixedCell{22pt}{52.99}{28pt}{} & \GACFixedCell{22pt}{94.73}{28pt}{} & \GACFixedCell{22pt}{53.01}{28pt}{} & \GACFixedCell{22pt}{52.69}{28pt}{} & \GACFixedCell{22pt}{39.78}{28pt}{}\\
\GACPDFRowFill
 & \GACHTMLRowFill\hspace{0.8em}+ GAC & \GACFixedCell{22pt}{71.07}{28pt}{+12.35} & \GACFixedCell{22pt}{99.43}{28pt}{-0.01} & \GACFixedCell{22pt}{64.88}{28pt}{+11.89} & \GACFixedCell{22pt}{93.45}{28pt}{-1.28} & \GACFixedCell{22pt}{64.78}{28pt}{+11.77} & \GACFixedCell{22pt}{55.75}{28pt}{+3.06} & \GACFixedCell{22pt}{45.12}{28pt}{+5.34}\\
 & DiffusionDrive & \GACFixedCell{22pt}{43.24}{28pt}{} & \GACFixedCell{22pt}{95.48}{28pt}{} & \GACFixedCell{22pt}{35.05}{28pt}{} & \GACFixedCell{22pt}{90.24}{28pt}{} & \GACFixedCell{22pt}{34.37}{28pt}{} & \GACFixedCell{22pt}{41.38}{28pt}{} & \GACFixedCell{22pt}{26.99}{28pt}{}\\
\GACPDFRowFill
 & \GACHTMLRowFill\hspace{0.8em}+ GAC & \GACFixedCell{22pt}{59.48}{28pt}{+16.24} & \GACFixedCell{22pt}{97.03}{28pt}{+1.55} & \GACFixedCell{22pt}{52.60}{28pt}{+17.55} & \GACFixedCell{22pt}{90.02}{28pt}{-0.22} & \GACFixedCell{22pt}{50.99}{28pt}{+16.62} & \GACFixedCell{22pt}{49.07}{28pt}{+7.69} & \GACFixedCell{22pt}{37.28}{28pt}{+10.29}\\
\bottomrule
\GACEndTabularStar
\end{table*}

%% file: tables/Table7.tex
\begin{table*}[tp]
\centering
\caption{Closed-loop performance by difficulty on HUGSIM-160. Each difficulty level contains 40 scenarios. Shading marks GAC; purple values show score changes from Original.}
\label{tab:closed_loop_difficulties}
\GACTableStyle
\setlength{\tabcolsep}{2pt}
\GACBeginTabularStar{\linewidth}{@{}>{\raggedright\arraybackslash}m{43pt}>{\raggedright\arraybackslash}m{56pt}@{\extracolsep{\fill}}*{7}{l}@{}}{@{}ll*{7}{c}@{}}
\toprule
\textbf{Difficulty} & \textbf{Method} & \GACFixedCell{22pt}{\GACMetricHeader{NC}{NC~↑}}{28pt}{} & \GACFixedCell{22pt}{\GACMetricHeader{DAC}{DAC~↑}}{28pt}{} & \GACFixedCell{22pt}{\GACMetricHeader{TTC}{TTC~↑}}{28pt}{} & \GACFixedCell{22pt}{\GACMetricHeader{Comfort}{Comfort~↑}}{28pt}{} & \GACFixedCell{22pt}{\GACMetricHeader{PDMS}{PDMS~↑}}{28pt}{} & \GACFixedCell{22pt}{\GACMetricHeader{RC}{RC~↑}}{28pt}{} & \GACFixedCell{22pt}{\GACMetricHeader[72pt]{HD-Score}{HD-Score~↑}}{28pt}{}\\
\midrule
Easy & iDriveVLA & \GACFixedCell{22pt}{93.18}{28pt}{} & \GACFixedCell{22pt}{96.86}{28pt}{} & \GACFixedCell{22pt}{90.25}{28pt}{} & \GACFixedCell{22pt}{99.34}{28pt}{} & \GACFixedCell{22pt}{87.86}{28pt}{} & \GACFixedCell{22pt}{87.54}{28pt}{} & \GACFixedCell{22pt}{78.32}{28pt}{}\\
\GACPDFRowFill
 & \GACHTMLRowFill\hspace{0.8em}+ GAC & \GACFixedCell{22pt}{93.77}{28pt}{+0.59} & \GACFixedCell{22pt}{97.04}{28pt}{+0.18} & \GACFixedCell{22pt}{91.23}{28pt}{+0.98} & \GACFixedCell{22pt}{99.32}{28pt}{-0.02} & \GACFixedCell{22pt}{88.78}{28pt}{+0.92} & \GACFixedCell{22pt}{88.42}{28pt}{+0.88} & \GACFixedCell{22pt}{79.85}{28pt}{+1.53}\\
 & DrivoR & \GACFixedCell{22pt}{91.60}{28pt}{} & \GACFixedCell{22pt}{98.44}{28pt}{} & \GACFixedCell{22pt}{89.36}{28pt}{} & \GACFixedCell{22pt}{93.66}{28pt}{} & \GACFixedCell{22pt}{86.91}{28pt}{} & \GACFixedCell{22pt}{89.91}{28pt}{} & \GACFixedCell{22pt}{80.78}{28pt}{}\\
\GACPDFRowFill
 & \GACHTMLRowFill\hspace{0.8em}+ GAC & \GACFixedCell{22pt}{92.49}{28pt}{+0.89} & \GACFixedCell{22pt}{98.57}{28pt}{+0.13} & \GACFixedCell{22pt}{90.31}{28pt}{+0.95} & \GACFixedCell{22pt}{94.36}{28pt}{+0.70} & \GACFixedCell{22pt}{88.28}{28pt}{+1.37} & \GACFixedCell{22pt}{91.42}{28pt}{+1.51} & \GACFixedCell{22pt}{83.15}{28pt}{+2.37}\\
 & DiffusionDrive & \GACFixedCell{22pt}{83.72}{28pt}{} & \GACFixedCell{22pt}{97.05}{28pt}{} & \GACFixedCell{22pt}{80.92}{28pt}{} & \GACFixedCell{22pt}{92.98}{28pt}{} & \GACFixedCell{22pt}{77.54}{28pt}{} & \GACFixedCell{22pt}{78.50}{28pt}{} & \GACFixedCell{22pt}{68.65}{28pt}{}\\
\GACPDFRowFill
 & \GACHTMLRowFill\hspace{0.8em}+ GAC & \GACFixedCell{22pt}{86.87}{28pt}{+3.15} & \GACFixedCell{22pt}{98.01}{28pt}{+0.96} & \GACFixedCell{22pt}{85.08}{28pt}{+4.16} & \GACFixedCell{22pt}{92.84}{28pt}{-0.14} & \GACFixedCell{22pt}{81.58}{28pt}{+4.04} & \GACFixedCell{22pt}{85.34}{28pt}{+6.84} & \GACFixedCell{22pt}{75.42}{28pt}{+6.77}\\
\midrule
Medium & iDriveVLA & \GACFixedCell{22pt}{82.20}{28pt}{} & \GACFixedCell{22pt}{93.19}{28pt}{} & \GACFixedCell{22pt}{79.78}{28pt}{} & \GACFixedCell{22pt}{99.34}{28pt}{} & \GACFixedCell{22pt}{76.15}{28pt}{} & \GACFixedCell{22pt}{50.04}{28pt}{} & \GACFixedCell{22pt}{43.59}{28pt}{}\\
\GACPDFRowFill
 & \GACHTMLRowFill\hspace{0.8em}+ GAC & \GACFixedCell{22pt}{85.76}{28pt}{+3.56} & \GACFixedCell{22pt}{94.02}{28pt}{+0.83} & \GACFixedCell{22pt}{84.06}{28pt}{+4.28} & \GACFixedCell{22pt}{99.29}{28pt}{-0.05} & \GACFixedCell{22pt}{80.90}{28pt}{+4.75} & \GACFixedCell{22pt}{52.34}{28pt}{+2.30} & \GACFixedCell{22pt}{46.32}{28pt}{+2.73}\\
 & DrivoR & \GACFixedCell{22pt}{49.52}{28pt}{} & \GACFixedCell{22pt}{94.33}{28pt}{} & \GACFixedCell{22pt}{43.03}{28pt}{} & \GACFixedCell{22pt}{93.09}{28pt}{} & \GACFixedCell{22pt}{42.54}{28pt}{} & \GACFixedCell{22pt}{46.96}{28pt}{} & \GACFixedCell{22pt}{29.47}{28pt}{}\\
\GACPDFRowFill
 & \GACHTMLRowFill\hspace{0.8em}+ GAC & \GACFixedCell{22pt}{68.36}{28pt}{+18.84} & \GACFixedCell{22pt}{95.87}{28pt}{+1.54} & \GACFixedCell{22pt}{61.38}{28pt}{+18.35} & \GACFixedCell{22pt}{93.06}{28pt}{-0.03} & \GACFixedCell{22pt}{61.25}{28pt}{+18.71} & \GACFixedCell{22pt}{52.70}{28pt}{+5.74} & \GACFixedCell{22pt}{38.60}{28pt}{+9.13}\\
 & DiffusionDrive & \GACFixedCell{22pt}{37.83}{28pt}{} & \GACFixedCell{22pt}{90.06}{28pt}{} & \GACFixedCell{22pt}{29.18}{28pt}{} & \GACFixedCell{22pt}{91.45}{28pt}{} & \GACFixedCell{22pt}{28.27}{28pt}{} & \GACFixedCell{22pt}{34.79}{28pt}{} & \GACFixedCell{22pt}{17.70}{28pt}{}\\
\GACPDFRowFill
 & \GACHTMLRowFill\hspace{0.8em}+ GAC & \GACFixedCell{22pt}{64.34}{28pt}{+26.51} & \GACFixedCell{22pt}{92.45}{28pt}{+2.39} & \GACFixedCell{22pt}{53.45}{28pt}{+24.27} & \GACFixedCell{22pt}{92.13}{28pt}{+0.68} & \GACFixedCell{22pt}{52.69}{28pt}{+24.42} & \GACFixedCell{22pt}{48.31}{28pt}{+13.52} & \GACFixedCell{22pt}{35.95}{28pt}{+18.25}\\
\midrule
Hard & iDriveVLA & \GACFixedCell{22pt}{58.24}{28pt}{} & \GACFixedCell{22pt}{95.53}{28pt}{} & \GACFixedCell{22pt}{49.56}{28pt}{} & \GACFixedCell{22pt}{97.72}{28pt}{} & \GACFixedCell{22pt}{48.39}{28pt}{} & \GACFixedCell{22pt}{33.01}{28pt}{} & \GACFixedCell{22pt}{23.79}{28pt}{}\\
\GACPDFRowFill
 & \GACHTMLRowFill\hspace{0.8em}+ GAC & \GACFixedCell{22pt}{66.90}{28pt}{+8.66} & \GACFixedCell{22pt}{95.00}{28pt}{-0.53} & \GACFixedCell{22pt}{62.33}{28pt}{+12.77} & \GACFixedCell{22pt}{94.87}{28pt}{-2.85} & \GACFixedCell{22pt}{59.06}{28pt}{+10.67} & \GACFixedCell{22pt}{31.31}{28pt}{-1.70} & \GACFixedCell{22pt}{25.09}{28pt}{+1.30}\\
 & DrivoR & \GACFixedCell{22pt}{44.79}{28pt}{} & \GACFixedCell{22pt}{95.86}{28pt}{} & \GACFixedCell{22pt}{36.88}{28pt}{} & \GACFixedCell{22pt}{93.37}{28pt}{} & \GACFixedCell{22pt}{35.98}{28pt}{} & \GACFixedCell{22pt}{38.56}{28pt}{} & \GACFixedCell{22pt}{24.88}{28pt}{}\\
\GACPDFRowFill
 & \GACHTMLRowFill\hspace{0.8em}+ GAC & \GACFixedCell{22pt}{56.65}{28pt}{+11.86} & \GACFixedCell{22pt}{95.85}{28pt}{-0.01} & \GACFixedCell{22pt}{49.64}{28pt}{+12.76} & \GACFixedCell{22pt}{92.27}{28pt}{-1.10} & \GACFixedCell{22pt}{47.82}{28pt}{+11.84} & \GACFixedCell{22pt}{36.84}{28pt}{-1.72} & \GACFixedCell{22pt}{27.19}{28pt}{+2.31}\\
 & DiffusionDrive & \GACFixedCell{22pt}{35.42}{28pt}{} & \GACFixedCell{22pt}{91.59}{28pt}{} & \GACFixedCell{22pt}{26.78}{28pt}{} & \GACFixedCell{22pt}{87.35}{28pt}{} & \GACFixedCell{22pt}{23.16}{28pt}{} & \GACFixedCell{22pt}{27.85}{28pt}{} & \GACFixedCell{22pt}{11.84}{28pt}{}\\
\GACPDFRowFill
 & \GACHTMLRowFill\hspace{0.8em}+ GAC & \GACFixedCell{22pt}{54.51}{28pt}{+19.09} & \GACFixedCell{22pt}{93.87}{28pt}{+2.28} & \GACFixedCell{22pt}{44.54}{28pt}{+17.76} & \GACFixedCell{22pt}{87.56}{28pt}{+0.21} & \GACFixedCell{22pt}{40.92}{28pt}{+17.76} & \GACFixedCell{22pt}{33.86}{28pt}{+6.01} & \GACFixedCell{22pt}{21.93}{28pt}{+10.09}\\
\midrule
Extreme & iDriveVLA & \GACFixedCell{22pt}{30.47}{28pt}{} & \GACFixedCell{22pt}{95.88}{28pt}{} & \GACFixedCell{22pt}{23.85}{28pt}{} & \GACFixedCell{22pt}{96.64}{28pt}{} & \GACFixedCell{22pt}{24.40}{28pt}{} & \GACFixedCell{22pt}{24.86}{28pt}{} & \GACFixedCell{22pt}{11.39}{28pt}{}\\
\GACPDFRowFill
 & \GACHTMLRowFill\hspace{0.8em}+ GAC & \GACFixedCell{22pt}{39.86}{28pt}{+9.39} & \GACFixedCell{22pt}{95.82}{28pt}{-0.06} & \GACFixedCell{22pt}{31.51}{28pt}{+7.66} & \GACFixedCell{22pt}{94.13}{28pt}{-2.51} & \GACFixedCell{22pt}{32.05}{28pt}{+7.65} & \GACFixedCell{22pt}{23.61}{28pt}{-1.25} & \GACFixedCell{22pt}{12.41}{28pt}{+1.02}\\
 & DrivoR & \GACFixedCell{22pt}{31.55}{28pt}{} & \GACFixedCell{22pt}{95.21}{28pt}{} & \GACFixedCell{22pt}{24.65}{28pt}{} & \GACFixedCell{22pt}{95.10}{28pt}{} & \GACFixedCell{22pt}{24.81}{28pt}{} & \GACFixedCell{22pt}{30.15}{28pt}{} & \GACFixedCell{22pt}{14.53}{28pt}{}\\
\GACPDFRowFill
 & \GACHTMLRowFill\hspace{0.8em}+ GAC & \GACFixedCell{22pt}{40.38}{28pt}{+8.83} & \GACFixedCell{22pt}{94.53}{28pt}{-0.68} & \GACFixedCell{22pt}{33.78}{28pt}{+9.13} & \GACFixedCell{22pt}{92.61}{28pt}{-2.49} & \GACFixedCell{22pt}{33.10}{28pt}{+8.29} & \GACFixedCell{22pt}{30.65}{28pt}{+0.50} & \GACFixedCell{22pt}{17.53}{28pt}{+3.00}\\
 & DiffusionDrive & \GACFixedCell{22pt}{28.09}{28pt}{} & \GACFixedCell{22pt}{94.51}{28pt}{} & \GACFixedCell{22pt}{22.08}{28pt}{} & \GACFixedCell{22pt}{90.33}{28pt}{} & \GACFixedCell{22pt}{21.24}{28pt}{} & \GACFixedCell{22pt}{28.16}{28pt}{} & \GACFixedCell{22pt}{12.15}{28pt}{}\\
\GACPDFRowFill
 & \GACHTMLRowFill\hspace{0.8em}+ GAC & \GACFixedCell{22pt}{32.01}{28pt}{+3.92} & \GACFixedCell{22pt}{95.05}{28pt}{+0.54} & \GACFixedCell{22pt}{24.88}{28pt}{+2.80} & \GACFixedCell{22pt}{88.54}{28pt}{-1.79} & \GACFixedCell{22pt}{24.89}{28pt}{+3.65} & \GACFixedCell{22pt}{25.74}{28pt}{-2.42} & \GACFixedCell{22pt}{12.42}{28pt}{+0.27}\\
\bottomrule
\GACEndTabularStar
\end{table*}

%% file: tables/Table8.tex
\begin{table}[t]
\centering
\caption{Component scores by assessment frequency on HUGSIM-64. GAC performs assessment at 1 or 2\,Hz, while all planners operate at 4\,Hz. Purple values show score changes from Original.}
\label{tab:closed_loop_frequency_components}
\GACTableStyle
\setlength{\tabcolsep}{1.25pt}
\GACBeginTabularStar{\linewidth}{@{}>{\raggedright\arraybackslash}m{51pt}@{\extracolsep{\fill}}*{4}{l}@{}}{@{}l*{4}{c}@{}}
\toprule
\textbf{Method} & \GACFixedCell{18pt}{\GACMetricHeader{NC}{NC~↑}}{28pt}{} & \GACFixedCell{18pt}{\GACMetricHeader{DAC}{DAC~↑}}{24pt}{} & \GACFixedCell{18pt}{\GACMetricHeader{TTC}{TTC~↑}}{28pt}{} & \GACFixedCell{18pt}{\GACMetricHeader{Comf.}{Comf.~↑}}{24pt}{}\\
\midrule
iDriveVLA & \GACFixedCell{18pt}{62.79}{28pt}{} & \GACFixedCell{18pt}{96.29}{24pt}{} & \GACFixedCell{18pt}{57.34}{28pt}{} & \GACFixedCell{18pt}{98.19}{24pt}{}\\
\hspace{0.2em}+ GAC (1\,Hz) & \GACFixedCell{18pt}{65.96}{28pt}{+3.17} & \GACFixedCell{18pt}{96.46}{24pt}{+0.17} & \GACFixedCell{18pt}{60.71}{28pt}{+3.37} & \GACFixedCell{18pt}{97.39}{24pt}{-0.80}\\
\GACRowColor
\hspace{0.2em}+ GAC (2\,Hz) & \GACFixedCell{18pt}{67.63}{28pt}{+4.84} & \GACFixedCell{18pt}{96.56}{24pt}{+0.27} & \GACFixedCell{18pt}{63.15}{28pt}{+5.81} & \GACFixedCell{18pt}{96.90}{24pt}{-1.29}\\
\midrule
DrivoR & \GACFixedCell{18pt}{55.01}{28pt}{} & \GACFixedCell{18pt}{96.07}{24pt}{} & \GACFixedCell{18pt}{49.82}{28pt}{} & \GACFixedCell{18pt}{94.68}{24pt}{}\\
\hspace{0.2em}+ GAC (1\,Hz) & \GACFixedCell{18pt}{59.75}{28pt}{+4.74} & \GACFixedCell{18pt}{95.74}{24pt}{-0.33} & \GACFixedCell{18pt}{54.57}{28pt}{+4.75} & \GACFixedCell{18pt}{93.99}{24pt}{-0.69}\\
\GACRowColor
\hspace{0.2em}+ GAC (2\,Hz) & \GACFixedCell{18pt}{65.15}{28pt}{+10.14} & \GACFixedCell{18pt}{96.40}{24pt}{+0.33} & \GACFixedCell{18pt}{60.44}{28pt}{+10.62} & \GACFixedCell{18pt}{93.39}{24pt}{-1.29}\\
\midrule
DiffusionDrive & \GACFixedCell{18pt}{43.52}{28pt}{} & \GACFixedCell{18pt}{93.68}{24pt}{} & \GACFixedCell{18pt}{37.54}{28pt}{} & \GACFixedCell{18pt}{91.33}{24pt}{}\\
\hspace{0.2em}+ GAC (1\,Hz) & \GACFixedCell{18pt}{54.51}{28pt}{+10.99} & \GACFixedCell{18pt}{94.76}{24pt}{+1.08} & \GACFixedCell{18pt}{49.23}{28pt}{+11.69} & \GACFixedCell{18pt}{90.28}{24pt}{-1.05}\\
\GACRowColor
\hspace{0.2em}+ GAC (2\,Hz) & \GACFixedCell{18pt}{57.97}{28pt}{+14.45} & \GACFixedCell{18pt}{94.86}{24pt}{+1.18} & \GACFixedCell{18pt}{52.12}{28pt}{+14.58} & \GACFixedCell{18pt}{91.69}{24pt}{+0.36}\\
\bottomrule
\GACEndTabularStar
\end{table}

%% file: tables/Table9.tex
% Keep this table with the compact reasoning cases on one float page.
\iflatexml
\begin{table*}
\else
\newsavebox{\GACOpenLoopTableBox}
\begin{lrbox}{\GACOpenLoopTableBox}
\begin{minipage}{\textwidth}
\fi
\captionsetup{type=table,skip=4pt}
\centering\footnotesize
\setlength{\tabcolsep}{5pt}
\caption{Open-loop performance on the Intermediate and High-quality groups of NAVSIM v1. Values are means over three sampled sets per planner and group. Subscripts indicate $\pm$ twice the sample standard deviation of the three set-level means.}
\label{tab:sampled_variability}
\begin{tabular}{lll*{6}{c}}
\toprule
\textbf{Planner} & \textbf{Group} & \textbf{Selection} & \textbf{NC}$\uparrow$ & \textbf{DAC}$\uparrow$ & \textbf{TTC}$\uparrow$ & \textbf{EP}$\uparrow$ & \textbf{NC$\times$DAC}$\uparrow$ & \textbf{PDMS}$\uparrow$\\
\midrule
iDriveVLA & Intermediate & Original & $99.33_{\pm1.53}$ & $100.00_{\pm0.00}$ & $75.00_{\pm7.21}$ & $72.94_{\pm2.16}$ & $99.33_{\pm1.53}$ & $78.02_{\pm2.55}$\\
\GACPDFSampleRowFill &  & \GACHTMLRowFill GAC & $99.67_{\pm1.15}$ & $99.33_{\pm1.15}$ & $87.00_{\pm5.29}$ & $69.80_{\pm2.66}$ & $99.00_{\pm0.00}$ & $81.56_{\pm1.35}$\\
 & High-quality & Original & $100.00_{\pm0.00}$ & $100.00_{\pm0.00}$ & $100.00_{\pm0.00}$ & $96.19_{\pm1.39}$ & $100.00_{\pm0.00}$ & $98.41_{\pm0.58}$\\
\GACPDFSampleRowFill &  & \GACHTMLRowFill GAC & $99.67_{\pm1.15}$ & $100.00_{\pm0.00}$ & $99.00_{\pm0.00}$ & $92.07_{\pm1.34}$ & $99.67_{\pm1.15}$ & $96.22_{\pm0.68}$\\
\midrule
DrivoR & Intermediate & Original & $99.50_{\pm0.00}$ & $100.00_{\pm0.00}$ & $85.33_{\pm9.45}$ & $69.18_{\pm3.16}$ & $99.50_{\pm0.00}$ & $80.82_{\pm2.72}$\\
\GACPDFSampleRowFill &  & \GACHTMLRowFill GAC & $99.83_{\pm0.58}$ & $98.67_{\pm3.06}$ & $90.33_{\pm3.06}$ & $66.60_{\pm2.45}$ & $98.50_{\pm3.00}$ & $81.22_{\pm2.03}$\\
 & High-quality & Original & $100.00_{\pm0.00}$ & $100.00_{\pm0.00}$ & $100.00_{\pm0.00}$ & $95.06_{\pm1.27}$ & $100.00_{\pm0.00}$ & $97.94_{\pm0.53}$\\
\GACPDFSampleRowFill &  & \GACHTMLRowFill GAC & $99.67_{\pm1.15}$ & $99.00_{\pm2.00}$ & $99.67_{\pm1.15}$ & $90.44_{\pm3.08}$ & $98.67_{\pm1.15}$ & $95.10_{\pm1.22}$\\
\midrule
DiffusionDrive & Intermediate & Original & $99.67_{\pm1.15}$ & $100.00_{\pm0.00}$ & $83.67_{\pm4.62}$ & $71.86_{\pm1.36}$ & $99.67_{\pm1.15}$ & $81.31_{\pm1.20}$\\
\GACPDFSampleRowFill &  & \GACHTMLRowFill GAC & $99.67_{\pm1.15}$ & $98.67_{\pm2.31}$ & $89.67_{\pm4.62}$ & $69.24_{\pm1.22}$ & $98.33_{\pm1.15}$ & $81.94_{\pm2.09}$\\
 & High-quality & Original & $100.00_{\pm0.00}$ & $100.00_{\pm0.00}$ & $100.00_{\pm0.00}$ & $91.84_{\pm0.57}$ & $100.00_{\pm0.00}$ & $96.60_{\pm0.24}$\\
\GACPDFSampleRowFill &  & \GACHTMLRowFill GAC & $100.00_{\pm0.00}$ & $99.67_{\pm1.15}$ & $99.33_{\pm1.15}$ & $89.48_{\pm0.55}$ & $99.67_{\pm1.15}$ & $95.14_{\pm0.33}$\\
\bottomrule
\end{tabular}
\iflatexml
\end{table*}
\else
\end{minipage}
\end{lrbox}
\fi

%% file: sections/supplementary_temporal_analysis.tex
\subsection{Temporal Analysis of Safety Degradation}\label{app:temporal_analysis}
Among the 1,800 Intermediate and High-quality evaluation entries, 260 show a decrease in PDMS after GAC selection. In 234 cases (90.0\%), only EP decreases, suggesting that most performance reductions are associated with reduced progress rather than safety-related degradation. We further examine the remaining 26 cases with declines in NC, DAC, or TTC, including 15 Intermediate and 11 High-quality entries. We replay the original and GAC-selected trajectories using the NAVSIM evaluator, confirm that their scores match the original evaluations, and examine when safety-related violations first occur along the four-second trajectories. All 26 GAC-selected trajectories retain full NC, DAC, and TTC scores through the first second, with the first violations detected at 1.7--4.0 seconds for Intermediate entries and 2.0--3.9 seconds for High-quality entries. In the 2\,Hz closed-loop configuration, GAC reassesses planner selections every 0.5 seconds using updated observations, enabling decisions to be updated before the full trajectory is executed.

%% file: sections/reasoning_cases.tex
\begin{figure*}[p]
\iflatexml\else
\usebox{\GACOpenLoopTableBox}
\par\vspace{20pt}
\fi
\centering
\captionsetup{skip=4pt}
\iflatexml\else
\begingroup%
\phantomsubcaption\label{fig:reasoning_truck}%
\phantomsubcaption\label{fig:reasoning_acceptance}%
\endgroup%
\fi%
\iflatexml
\GACHTMLFullWidthGraphic[alt={(a) Rejection and reselection. DrivoR candidate 4 is rejected due to a potential rear-end conflict with a lead truck, and candidate 43 is selected, increasing PDMS from 0 to 96.04. (b) Acceptance. Candidate 51 is retained with PDMS 100. Each case includes historical and current front views, the candidate set, and the recorded critic reasoning.}]{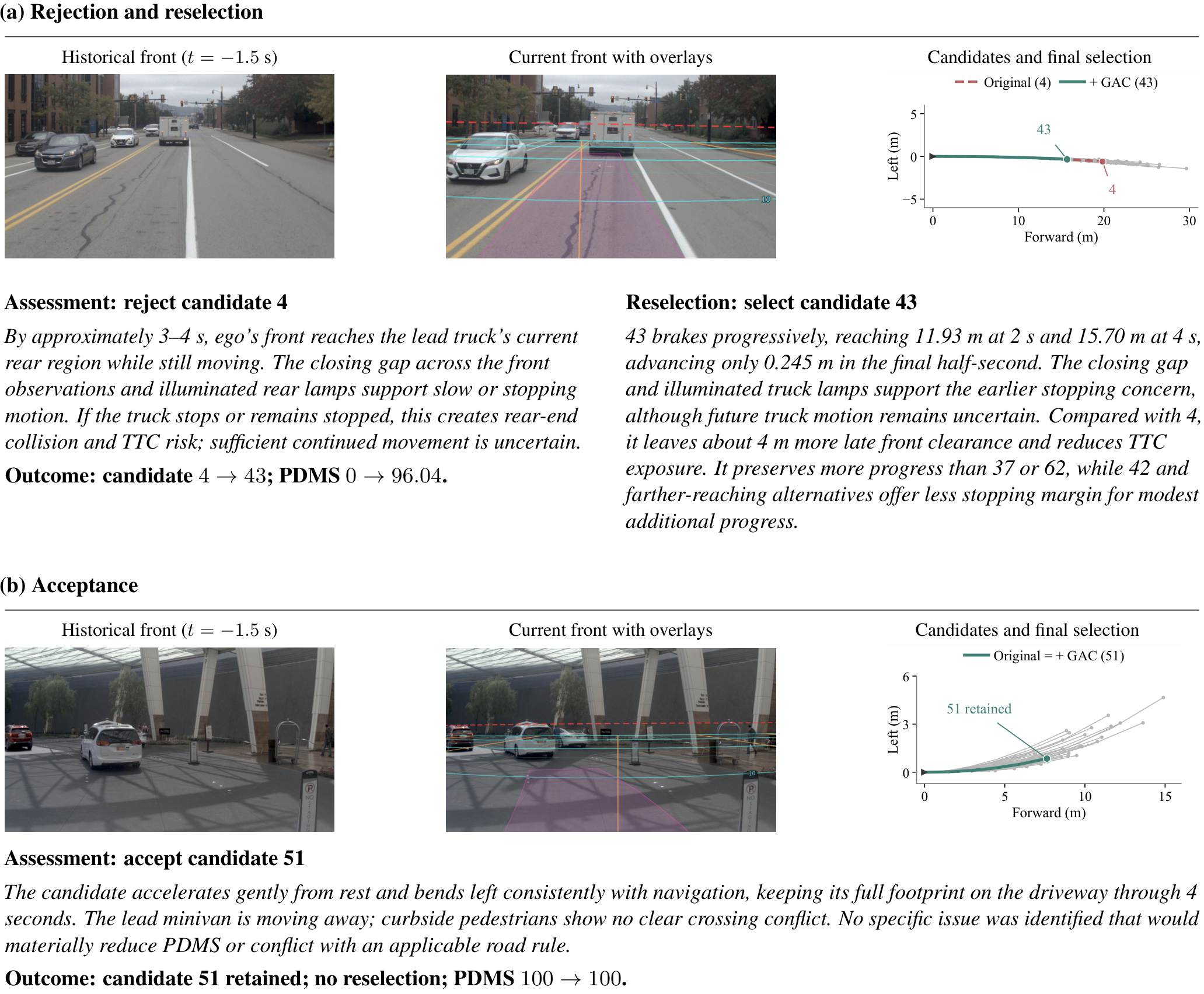}
\else
\includegraphics[width=\linewidth,alt={(a) Rejection and reselection. DrivoR candidate 4 is rejected due to a potential rear-end conflict with a lead truck, and candidate 43 is selected, increasing PDMS from 0 to 96.04. (b) Acceptance. Candidate 51 is retained with PDMS 100. Each case includes historical and current front views, the candidate set, and the recorded critic reasoning.}]{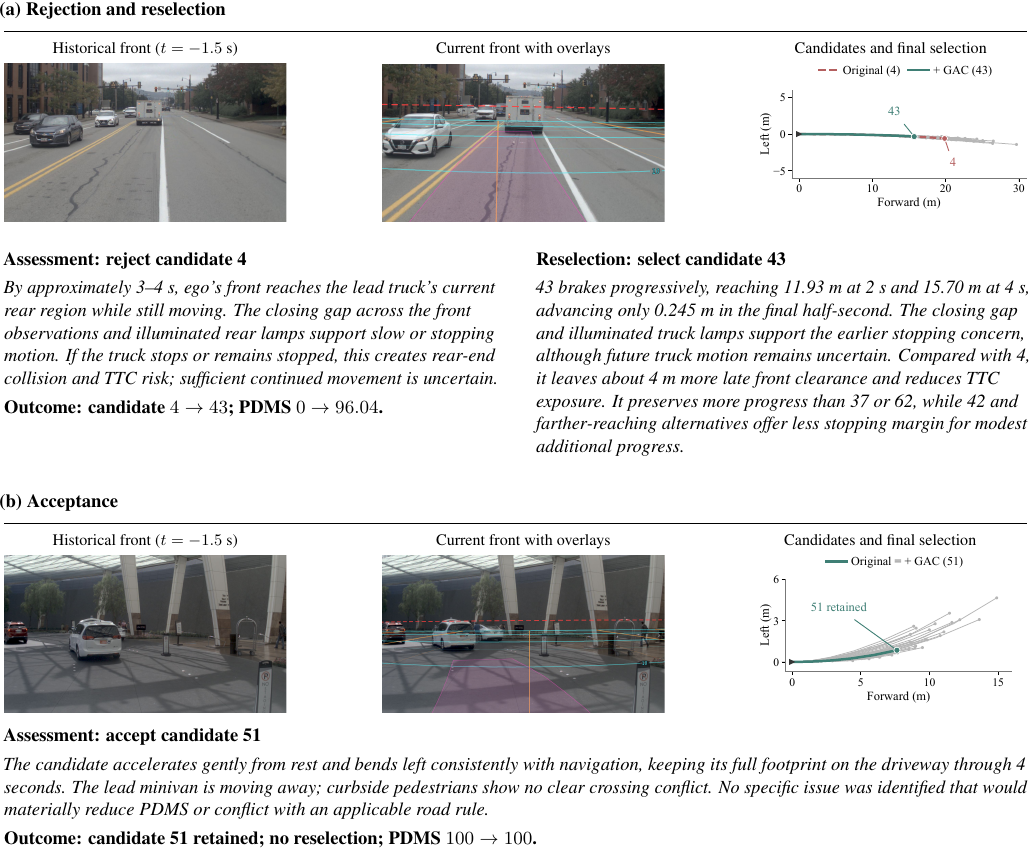}
\fi%
\par\vspace{2pt}
\caption{Assessment and reselection examples from DrivoR on NAVSIM v1. The left and middle panels show historical and annotated current front views supplied to the critic. Overlays depict the original candidate. The right panel shows all 64 candidates in the ego frame, with original and final selections highlighted and dots marking $4$\,s endpoints. The text reproduces the critic's recorded reasons verbatim.}
\label{fig:reasoning_cases}
\end{figure*}

Figure~\ref{fig:reasoning_cases} presents two DrivoR examples from our NAVSIM v1 evaluation, illustrating rejection followed by reselection and acceptance without reselection. In \figpanelref{fig:reasoning_cases}{fig:reasoning_truck}{a}, the decreasing distance to the lead truck and the original trajectory's forward extent indicate a potential rear-end collision risk. GAC selects a trajectory with greater stopping margin while maintaining forward progress, improving PDMS from 0 to 96.04. In \figpanelref{fig:reasoning_cases}{fig:reasoning_acceptance}{b}, the critic identifies no clear conflict with the departing minivan or pedestrians near the curb and retains the original selection, preserving a PDMS of 100.

%% file: sections/closed_loop_cases.tex
Additional qualitative examples of closed-loop driving on HUGSIM are provided in Figure~\ref{fig:closed_loop_multiview_v2}, illustrating how GAC improves driving behavior through trajectory selection in scenarios involving braking, lateral clearance, and turning.

\input{figures/Figure9_layout}

%% file: figures/Figure9_layout.tex
\begin{figure*}[p]
\centering
\captionsetup{skip=5pt}
\iflatexml\else
\begingroup%
\phantomsubcaption\label{fig:v2_hug_boundary}%
\phantomsubcaption\label{fig:v2_hug_stop}%
\phantomsubcaption\label{fig:v2_hug_navigation}%
\phantomsubcaption\label{fig:v2_hug_lateral}%
\endgroup%
\fi%
\iflatexml
\GACHTMLFullWidthGraphic{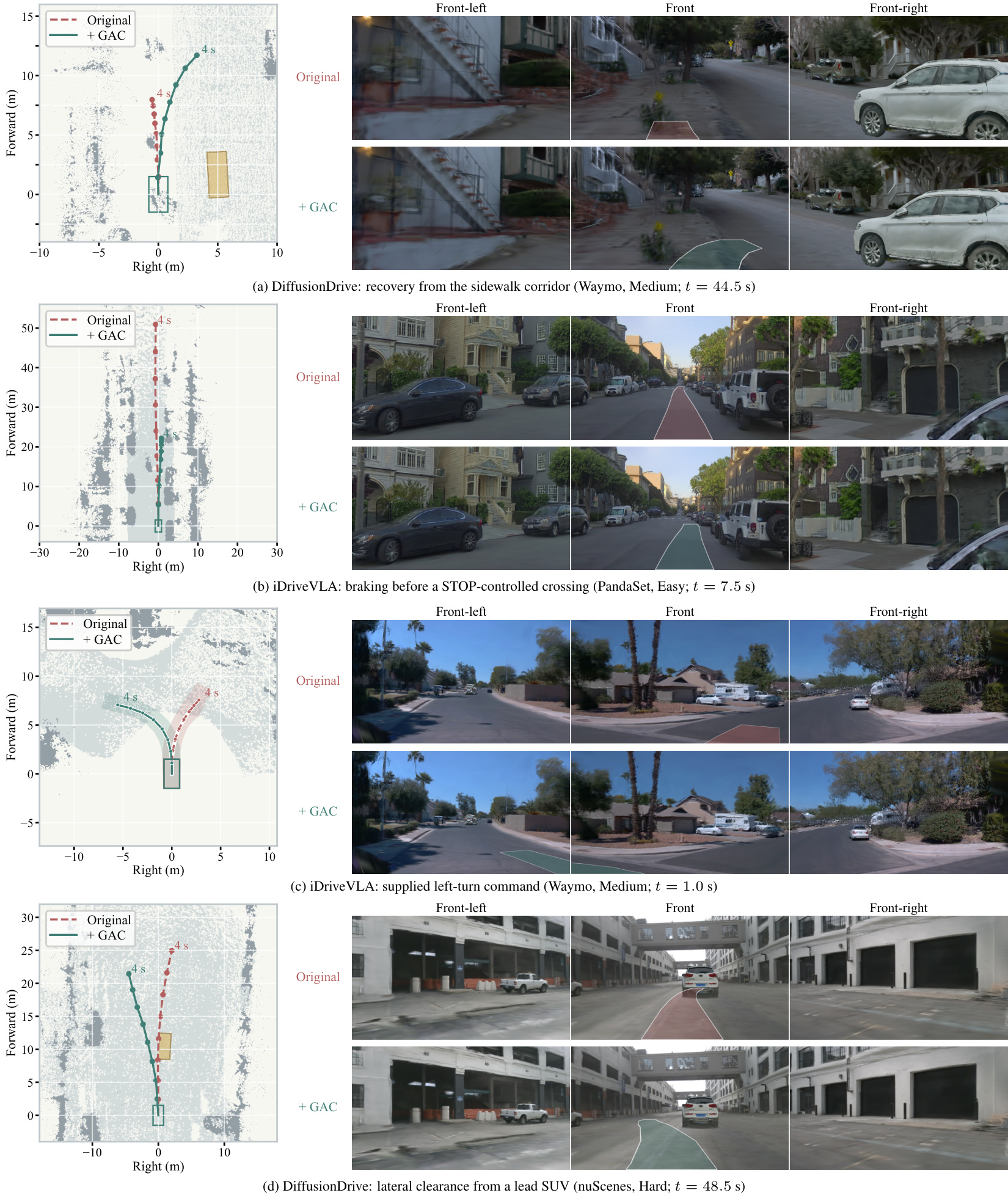}%
\else
\includegraphics[width=\linewidth]{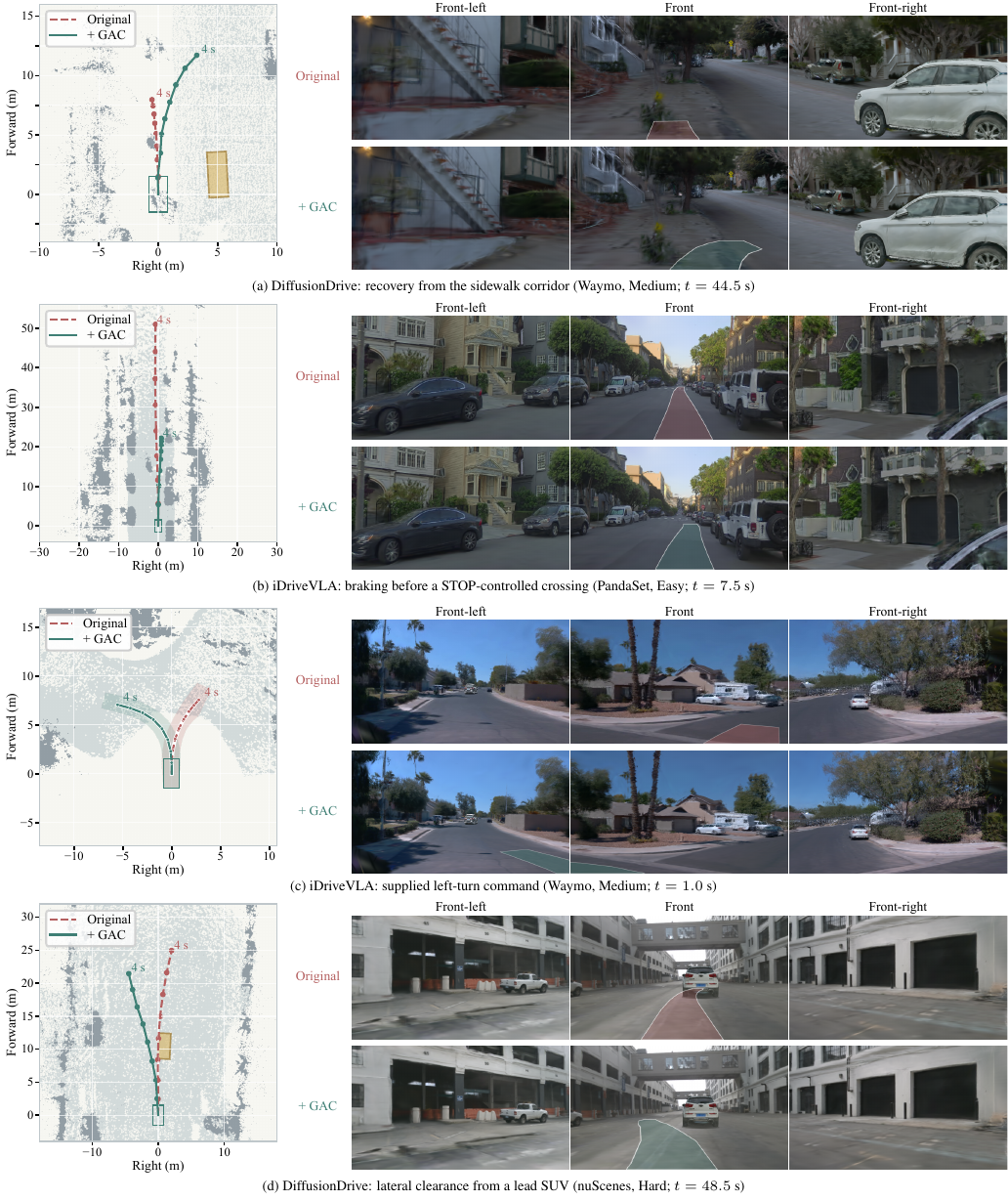}%
\fi%
\caption{Qualitative comparisons of Original and GAC in closed-loop driving on HUGSIM. Each case presents front-left, front, front-right, and BEV views, with Original above GAC. The red paths indicate trajectories selected by Original, while the green paths indicate those selected by GAC. All trajectories span a four-second horizon.}
\label{fig:closed_loop_multiview_v2}\label{fig:closed_loop_additional_case}
\end{figure*}

%% file: sections/open_loop_cases.tex
Additional qualitative examples of failure recovery on NAVSIM v1 are provided in Figure~\ref{fig:navsim_cases_v2}, illustrating how GAC improves collision avoidance and drivable-area compliance through trajectory reselection across diverse driving scenarios.

\input{figures/Figure10_layout}

%% file: figures/Figure10_layout.tex
\begin{figure*}[p]
\centering
\iflatexml\else
\begingroup%
\phantomsubcaption\label{fig:recovery_construction}%
\phantomsubcaption\label{fig:additional_zebra}%
\phantomsubcaption\label{fig:additional_oncoming}%
\phantomsubcaption\label{fig:recovery_truck}\label{fig:additional_truck}%
\phantomsubcaption\label{fig:recovery_separator}%
\phantomsubcaption\label{fig:additional_fountain}%
\phantomsubcaption\label{fig:recovery_parked_van}%
\phantomsubcaption\label{fig:recovery_splitter}%
\phantomsubcaption\label{fig:recovery_left_van}%
\endgroup%
\fi%
\iflatexml
\GACHTMLFullWidthGraphic{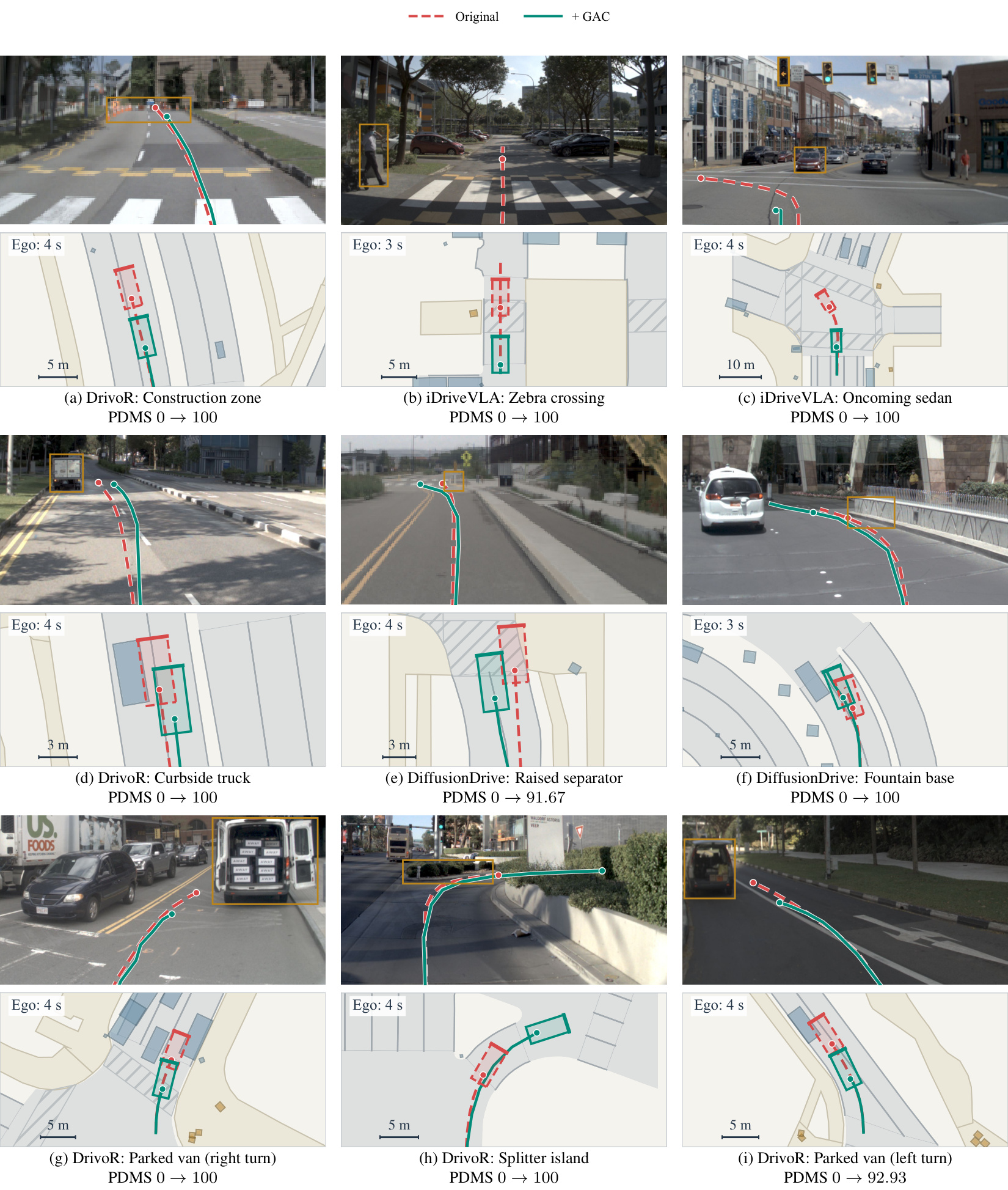}%
\else
\includegraphics[width=\linewidth]{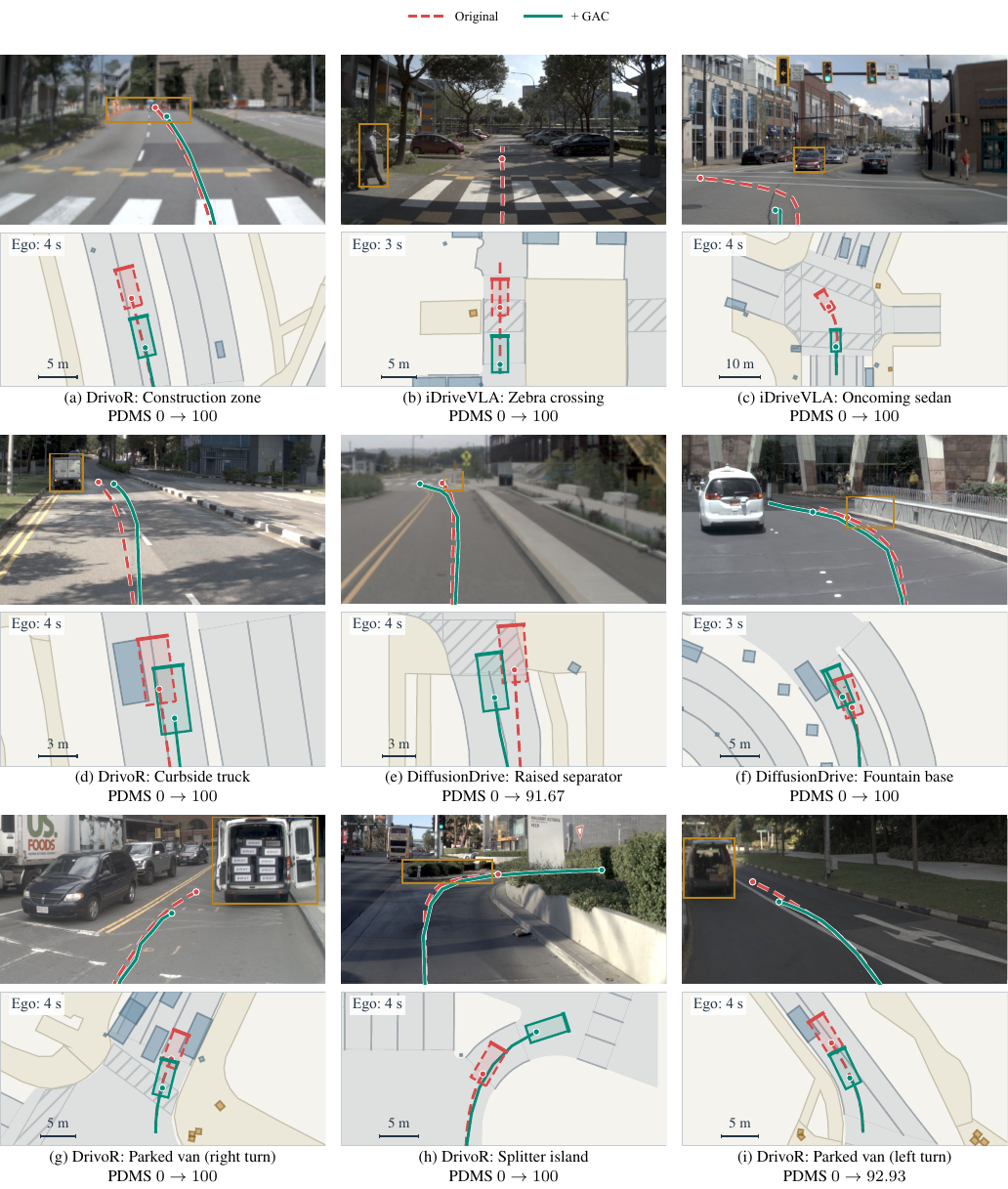}%
\fi%
\caption{Qualitative examples of open-loop failure recovery with GAC on NAVSIM v1. Each panel pairs a current front view with a BEV and reports the Original and GAC scores. Images and actor positions correspond to $t=0$, while trajectories span four seconds. Dots and ego footprints mark $t=3$\,s in (b) and (f), and $t=4$\,s otherwise. Scale bars are in metres.}
\label{fig:navsim_cases_v2}\label{fig:experiment_cases}
\label{fig:additional_recovery_cases}
\end{figure*}